\documentclass[10pt,twocolumn]{article}

\usepackage[utf8]{inputenc}
\usepackage[T1]{fontenc}
\usepackage{times}
\usepackage[margin=0.75in]{geometry}
\usepackage{graphicx}
\usepackage{amsmath,amssymb}
\usepackage{booktabs}
\usepackage{multirow}
\usepackage{caption}
\usepackage{subcaption}
\usepackage{hyperref}
\usepackage{xcolor}
\usepackage{enumitem}
\usepackage{array}
\usepackage{tabularx}
\usepackage{adjustbox}
\usepackage{float}
\usepackage{placeins}
\usepackage{algorithm}
\usepackage{algorithmic}
\usepackage{balance}
\usepackage{natbib}

\newcommand{\bench}{\textsc{DocForge-Bench}}

\newcommand{\eg}{\textit{e.g.}}

\title{\bench: A Comprehensive Benchmark for\\Document Forgery Detection and Analysis}

\author{%
  Zengqi Zhao$^{1,*}$ \quad
  Weidi Xia$^{2,*}$ \quad
  En Wei$^{3,*}$ \quad
  Yan Zhang$^{4,*}$ \quad
  Jane Mo$^{5,*}$ \\[4pt]
  Tiannan Zhang$^{6,*}$ \quad
  Yuanqin Dai$^{4,*}$ \quad
  Zexi Chen$^{7,*}$ \quad
  Yiran Tao$^{8,*}$ \quad
  Simiao Ren$^{4,\dagger}$
  \\[8pt]
  {\small $^{1}$University of North Carolina at Chapel Hill \quad
    $^{2}$University of California, Irvine \quad
    $^{3}$Washington University in St.\ Louis}\\[2pt]
  {\small $^{4}$Scam.ai \quad
    $^{5}$Duke University \quad
    $^{6}$University of California, Davis}\\[2pt]
  {\small $^{7}$New York University \quad
    $^{8}$Georgetown University}\\[4pt]
  {\small $^{*}$Equal contribution; order selected randomly.\quad
    $^{\dagger}$Corresponding author: \texttt{benren@scam.ai}}
}

\date{}

\begin{document}
\maketitle
{\renewcommand{\thefootnote}{}\footnote{Authorship of this paper does not imply employment by or affiliation with Scam.AI (Reality Inc.); some authors are independent researchers who collaborated on this work.}\addtocounter{footnote}{-1}}

\begin{abstract}
We present \bench{}, the first unified zero-shot benchmark for document forgery detection, evaluating 14 methods across eight datasets spanning text tampering, receipt forgery, and identity document manipulation. Unlike fine-tuning-oriented evaluations such as ForensicHub~\citep{du2025forensichub}, \bench{} applies all methods with their published pretrained weights and \emph{no domain adaptation}---a deliberate design choice that reflects the realistic deployment scenario where practitioners lack labeled document training data. Our central finding is a pervasive \emph{calibration failure} invisible under single-threshold protocols: methods achieve moderate Pixel-AUC ($\geq$0.76) yet near-zero Pixel-F1. This AUC--F1 gap is not a discrimination failure but a score-distribution shift: tampered regions occupy only 0.27--4.17\% of pixels in document images---an order of magnitude less than in natural image benchmarks---making the standard $\tau{=}0.5$ threshold catastrophically miscalibrated. Oracle-F1 is 2--10$\times$ higher than fixed-threshold Pixel-F1, confirming that calibration, not representation, is the bottleneck. A controlled calibration experiment validates this: adapting a single threshold on $N{=}10$ domain images recovers 39--55\% of the Oracle-F1 gap in representative evaluated high-AUC cases, demonstrating that threshold adaptation---not retraining---is the key missing step for practical deployment. Overall, \emph{no evaluated method works reliably out-of-the-box on diverse document types}, underscoring that document forgery detection remains an unsolved problem. We further note that all eight datasets predate the era of generative AI editing; benchmarks covering diffusion- and LLM-based document forgeries represent a critical open gap on the modern attack surface.
\end{abstract}

\section{Introduction}
\label{sec:intro}

Document forgery detectors trained on natural images fail in a specific and diagnostic way when applied to documents: they correctly rank forged pixels above authentic ones---achieving AUC $\geq$0.76---yet produce near-zero pixel-F1 at any fixed operating threshold. This is not a discrimination failure but a \emph{calibration failure}: score distributions shift systematically away from the standard $\tau{=}0.5$ decision boundary in the document domain. We demonstrate this failure empirically across 14 methods and 8 document datasets, establish it as the dominant bottleneck in zero-shot document forensics, and show that it is quantitatively explained by the tampered-pixel base rate---documents typically have 0.3--4\% forged pixels, an order of magnitude below the 10--30\% assumed by detectors trained on natural image benchmarks.

The image forensics community has made substantial progress in detecting manipulations in natural photographs. Methods such as TruFor~\citep{guillaro2023trufor}, MVSS-Net~\citep{chen2021mvss}, and CAT-Net~\citep{kwon2022catnet} achieve strong performance on established benchmarks like CASIA~\citep{dong2013casia}, Columbia~\citep{hsu2006columbia}, and NIST Nimble Challenge~\citep{nist2016}. However, these benchmarks and methods are designed for natural scenes---photographs of people, objects, and landscapes---and do not adequately address the unique characteristics of document images.

Document forgery presents fundamentally different challenges compared to natural image manipulation:
\begin{enumerate}[nosep,leftmargin=*]
    \item \textbf{Structured content:} Documents have rigid layouts with text, tables, logos, and stamps arranged in predictable patterns. Forgeries target semantic content (changing a name, amount, or date) rather than visual plausibility.
    \item \textbf{High-resolution text:} Detecting character-level modifications requires fine-grained analysis at resolutions where individual glyphs are distinguishable---a regime where most general forensic methods lose sensitivity.
    \item \textbf{Extreme region imbalance:} Document forgeries typically modify a few characters or fields, leaving 95--99.7\% of pixels authentic. Natural image benchmarks assume 10--30\% tampered area~\citep{dong2013casia,hsu2006columbia,wen2016coverage}; document datasets range from 0.27\% (ReceiptForgery) to 4.17\% (FSTS-1.5k). This order-of-magnitude difference invalidates the standard $\tau{=}0.5$ decision threshold used by all published methods.
    \item \textbf{Diverse forgery types:} A comprehensive system must handle text replacement, receipt price manipulation, and face swap in identity documents---each leaving distinct forensic traces.
\end{enumerate}

Despite these differences, evaluations remain fragmented. Recent efforts like DocTamper~\citep{qu2023doctamper} address text-level document tampering at scale, and ForensicHub~\citep{du2025forensichub} (NeurIPS 2025) makes significant progress toward unified evaluation across deepfake, image manipulation, AIGC, and document domains. However, ForensicHub evaluates document methods under a fine-tuning protocol and reports only fixed-threshold F1---measuring adapted performance but obscuring whether methods generalise out-of-the-box, and masking the calibration failures that make practical deployment hard. No benchmark focuses on document forgery with zero-shot frozen evaluation, threshold-independent metrics, and coverage of practically important document types such as physical receipts and identity cards.

\paragraph{Contributions.} We address this gap with \bench{}, offering:
\begin{enumerate}[nosep,leftmargin=*]
    \item \textbf{Zero-shot document benchmark:} The first unified zero-shot benchmark specifically for document forgery detection, cataloging 20 methods and fully evaluating 14 with publicly available pretrained weights and \emph{no domain fine-tuning} across 8 datasets---covering text tampering, identity document forgery, and receipt manipulation. IMDLBenCo~\citep{imdlbenco2024} focuses on natural image forensics; ForensicHub~\citep{du2025forensichub} includes fine-tuned scenarios and does not cover the full document dataset spectrum evaluated here. Unlike both, \bench{} assesses all methods at frozen published weights, isolating true out-of-the-box generalisation in the document domain.
    \item \textbf{Calibration gap diagnosis:} We report Pixel-AUC and Oracle-F1 alongside fixed-threshold F1, exposing a pervasive calibration gap in which methods correctly rank tampered pixels (high AUC) but fail to identify a usable decision threshold (near-zero F1). While it is well-known in the segmentation literature that fixed-threshold F1 degrades under class imbalance~\citep{lipton2014thresholding,boyd2012area}, our work provides the first empirical characterisation of this effect across 14 methods in the document forensics domain, quantifies the specific base-rate mismatch (0.27--4.17\% tampered pixels vs.\ 10--30\% in natural image benchmarks), and demonstrates practical recovery via threshold adaptation---a failure mode that remains invisible under single-threshold protocols.
    \item \textbf{Broader method and dataset coverage:} Seven general forensic methods and seven document-specific methods (including ASCFormer and ADCD-Net absent from prior benchmarks) evaluated on four datasets not covered elsewhere: ReceiptForgery, MixTamper, FSTS-1.5k, and FantasyID.
    \item \textbf{Mechanistic explanation of the gap:} We show quantitatively that the AUC--F1 gap is driven by tampered-pixel base rates (0.27--4.17\%) that are between 3$\times$ and 100$\times$ lower than in natural image benchmarks, making $\tau{=}0.5$ catastrophically miscalibrated---and that this is correctable via threshold adaptation on a small domain sample, without retraining.
\end{enumerate}

\section{Related Work}
\label{sec:related}

\subsection{Image Manipulation Detection}

Image forgery detection has evolved from hand-crafted feature methods to deep learning approaches. Early work exploited statistical artifacts such as JPEG compression inconsistencies~\citep{farid2009digital}, Error Level Analysis (ELA), and noise pattern analysis. The shift to deep learning began with methods like ManTraNet~\citep{wu2019mantranet}, which learns to detect 385 manipulation types in an end-to-end fashion without forgery-specific supervision.

Recent methods leverage increasingly sophisticated architectures. MVSS-Net~\citep{chen2021mvss} introduces multi-view (RGB + noise) and multi-scale supervision for boundary-aware detection. CAT-Net~\citep{kwon2022catnet} traces JPEG compression artifacts through dual-stream (RGB + DCT) analysis. PSCC-Net~\citep{liu2022pscc} employs progressive spatio-channel correlation for coarse-to-fine localization. Transformer-based approaches, including ObjectFormer~\citep{wang2022objectformer} and IML-ViT~\citep{ma2023imlvit}, demonstrate that vision transformers can capture long-range forensic dependencies. TruFor~\citep{guillaro2023trufor} combines a learned noise fingerprint (Noiseprint++) with RGB features via cross-modal fusion, achieving state-of-the-art performance across multiple benchmarks.

\subsection{Document-Specific Forgery Detection}

Document forgery detection has received comparatively less attention from the deep learning community. Traditional approaches relied on font analysis, alignment checking, and compression artifact detection in scanned documents. The introduction of DocTamper~\citep{qu2023doctamper} marked a significant advance, providing the first large-scale dataset ($\sim$170K images) for text-level document tampering detection. The accompanying baseline model uses a SegFormer-based architecture trained specifically on document text manipulations.

More recent work has broadened the scope of document forensics. DTD~\citep{qu2023dtd} (CVPR 2023) uses a dual-stream ConvNeXt+Swin-V2 architecture with JPEG DCT inputs and jointly introduced the DocTamper dataset; FFDN~\citep{chen2024ffdn} (ECCV 2024) fuses ConvNeXt RGB features with a DWT frequency pyramid; CAFTB-Net~\citep{song2024caftbnet} (TOMM 2024) applies SegFormer-B5 for high-frequency branch encoding; and TIFDM~\citep{dong2024tifdm} (TCE 2024) applies a ResNet-50 decoder with forgery trace enhancement. ASCFormer~\citep{ascformer2024} introduced the RealTextManipulation (RTM) benchmark and a transformer segmentation model for real-world text tampering in scene and document images. The OSTF dataset~\citep{daf2025} and accompanying DAF model target online tampered scene-text images generated by nine forgery engines. ADCD-Net~\citep{adcdnet2025} achieves ICCV 2025 state of the art by combining JPEG DCT analysis with RGB features and OCR-derived character masks.

\textbf{VLM-based approaches.} Recent work fine-tunes large vision-language models for document forgery, including TextShield-R1~\citep{textshield2026} and LogicLens~\citep{logiclens2025}. These represent a promising direction but are outside the scope of \bench{}, which focuses on methods that produce dense pixel-level localization masks. VLM-based approaches typically produce natural language descriptions or coarse bounding regions rather than binary spatial masks; evaluating them under our pixel-F1/AUC protocol would require non-trivial post-processing that introduces design choices beyond our zero-shot evaluation scope~\citep{liang2025llmdoc}. Pixel-level forensic localization with VLMs remains an open problem.

Identity document verification has developed as a parallel line of work, with the MIDV dataset series~\citep{bulatov2020midv2020} providing video and image data of identity documents. SIDTD~\citep{sidtd2023} offers synthetic identity document tampering with controlled forgery operations. The FantasyID dataset~\citep{fantasyid2025} targets KYC-scenario identity document forgery with face swap and AI text replacement attacks. However, evaluation across these datasets remains fragmented, and no prior work unifies general image forensics and document-specific detection under a common protocol.

\subsection{AI-Generated Content Detection}

The emergence of generative adversarial networks (GANs) and diffusion models has created a new class of forgeries. CNNDetection~\citep{wang2020cnndetection} demonstrated that a classifier trained on ProGAN images generalizes to other GAN architectures. UnivFD~\citep{ojha2023univfd} showed that frozen CLIP features with a linear probe provide surprisingly strong cross-generator detection. DIRE~\citep{wang2023dire} exploits diffusion model reconstruction error as a detection signal. \citet{ren2026aigc} provide a complementary zero-shot evaluation of open-source AIGC detectors, and \citet{ren2025llmdeepfake} assess whether multimodal LLMs can serve as deepfake detectors out of the box---both sharing our frozen-evaluation methodology.

\subsection{Evaluation Metrics and Protocols}
\label{sec:related_metrics}

Evaluation of forgery detection methods spans three distinct granularities, and the choice among them substantially affects reported numbers and inter-study comparability.

\textbf{Pixel-level localization.} The dominant paradigm in image manipulation detection treats forgery localization as binary segmentation: each pixel is labeled tampered or authentic, and performance is measured against a pixel-level ground-truth mask. The two primary metrics are Pixel-F1 and Pixel-AUC. Pixel-F1 at a fixed threshold $\tau{=}0.5$ is the harmonic mean of pixel-level precision and recall; it directly reflects deployment performance without calibration. Pixel-AUC is the area under the ROC curve swept over all thresholds and measures how well the model \emph{ranks} tampered pixels above authentic ones, independent of calibration. A critical but underappreciated distinction is whether F1 is computed at a fixed threshold or at the per-image \emph{optimal} threshold: the latter can inflate scores by 20--30 percentage points over fixed-threshold evaluation~\citep{imdlbenco2024}. This inconsistency pervades the document forgery literature---DTD~\citep{qu2023dtd}, FFDN~\citep{chen2024ffdn}, and ADCD-Net~\citep{adcdnet2025} all report F1 at an optimized threshold, while ForensicHub~\citep{du2025forensichub} and IMDLBenCo~\citep{imdlbenco2024} standardize on $\tau{=}0.5$. Additional pixel-level metrics include IoU (Jaccard index; algebraically equivalent to $\mathrm{F1}/(2-\mathrm{F1})$), Matthews Correlation Coefficient (MCC; balances all four confusion-matrix cells and is more robust under class imbalance~\citep{guillaro2023trufor}), and Average Precision (area under the Precision-Recall curve~\citep{safire2025}).

\textbf{Instance-level detection.} When ground truth is expressed as \emph{bounding boxes} rather than pixel masks---as in Tampered-IC13, OSTF~\citep{daf2025}, and ReceiptForgery~\citep{receiptforgery2023}---evaluation follows the object detection tradition. A predicted box is a true positive if its box-IoU with a ground-truth annotation meets a threshold (commonly 0.5 or the COCO standard of 0.5:0.05:0.95). Average Precision (AP) is then the area under the precision-recall curve over all confidence thresholds; mAP averages AP across forgery classes or across the IoU sweep. OSTF introduces a structured $9{\times}9$ matrix of forgery-engine $\times$ test-set configurations and reports the mean F1 (mF) across all 81 settings to capture cross-engine generalization. Instance-level metrics are fundamentally incompatible with pixel-level metrics unless boxes are filled to binary masks or masks are reduced to their bounding rectangle, each conversion introducing information loss.

\textbf{Image-level detection.} For the binary question of whether an image has been tampered at all, the standard metric is image-level AUC-ROC, used by TruFor~\citep{guillaro2023trufor}, MVSS-Net~\citep{chen2021mvss}, and all AI-generated image detectors~\citep{wang2020cnndetection,ojha2023univfd,wang2023dire}. CNNDetection and UnivFD instead report image-level Average Precision (mAP) because it is insensitive to the fraction of fake images in the test set. Identity document benchmarks (FantasyID~\citep{fantasyid2025}, SIDTD~\citep{sidtd2023}) follow the ISO/IEC 30107-3 biometric standard, reporting APCER (attack error rate) and BPCER (bona fide error rate) rather than forensics metrics.

\textbf{Implications for this benchmark.} The fragmentation across metric conventions makes cross-paper comparison unreliable. \bench{} standardizes on two complementary metrics applied uniformly to all 14 methods: Pixel-F1 at $\tau{=}0.5$ (deployment-relevant localization) and Pixel-AUC (calibration-independent ranking). Reporting both jointly is the key diagnostic: a high AUC with near-zero F1 is the \emph{calibration gap} that is our central empirical finding, invisible under single-threshold protocols.

\subsection{Existing Benchmarks and Surveys}

Several surveys cover image forensics broadly~\citep{verdoliva2020media}, but document-specific surveys remain scarce. The NIST Media Forensics Challenge provides standardized evaluation for general image manipulation but requires data agreements and excludes document-specific tasks. IMDLBenCo~\citep{imdlbenco2024} offers a unified training/evaluation codebase for image manipulation detection but focuses on natural images. In adjacent domains, \citet{ren2025deepfakedetectors} evaluate deepfake detectors under realistic conditions and \citet{ren2026age} benchmark VLMs versus traditional architectures for age estimation---studies that, together with our work, form a broader effort to characterise out-of-the-box performance across diverse forensic and biometric tasks.

The most closely related work is ForensicHub~\citep{du2025forensichub} (NeurIPS 2025), which provides a broad unified framework spanning deepfake, image manipulation, AI-generated content, and document detection across 23 datasets and 42 methods. Within its document component, ForensicHub evaluates four methods (DTD, FFDN, CAFTB-Net, TIFDM) on five datasets under a \emph{fine-tuning} protocol---models are trained on DocTamper and then tested, measuring \emph{adapted} performance. It also reports only fixed-threshold F1 ($\tau{=}0.5$) and IoU, without AUC or threshold-independent metrics.

\bench{} differs from ForensicHub in three key ways.
\textbf{(1) Zero-shot frozen evaluation.} We evaluate all methods with their published pretrained weights and \emph{no domain fine-tuning}, isolating out-of-the-box generalization.  Fine-tuning on DocTamper substantially inflates in-domain numbers and obscures whether a method genuinely generalizes to diverse document types.
\textbf{(2) Calibration analysis.} By reporting Pixel-AUC alongside fixed-threshold F1, we reveal a pervasive calibration gap: methods can correctly rank tampered pixels (AUC $\geq$ 0.76) yet produce near-zero F1 because their score distributions are not calibrated to a useful operating threshold. This phenomenon---invisible under ForensicHub's single-threshold protocol---is one of our central findings.
\textbf{(3) Broader document coverage.} We evaluate seven document-specific methods (adding ASCFormer and ADCD-Net absent from ForensicHub), seven general forensic methods (adding ManTraNet and SAFIRE), and eight datasets (adding ReceiptForgery, MixTamper, FSTS-1.5k, and FantasyID). The latter two cover practically important document types---physical receipts and identity cards---not addressed by ForensicHub.

\smallskip\noindent\textbf{Terminology note.} Throughout this paper, \emph{zero-shot evaluation} refers to applying any method with its published pretrained weights and no domain adaptation; we use \emph{out-of-distribution (OOD)} to denote any (method, dataset) pair where the dataset was unseen during training.

\section{Document Forgery: Threat Models and Benchmark Coverage}
\label{sec:taxonomy}

Document forgery differs fundamentally from natural image manipulation in its threat model. Rather than splicing photographs for visual deception, document forgeries target \emph{semantic content}---altering a name, date, amount, or identity field to change the meaning of a legally or financially binding artifact. We organise our benchmark around three operationally distinct threat scenarios that motivate dataset selection.

\subsection{Text-Region Tampering}

The most prevalent operational threat involves modifying printed textual content within a document image. Forensically, this leaves artifacts in local font statistics, JPEG block boundaries, background texture consistency, and high-frequency edge profiles at character boundaries. Detection requires fine-grained analysis at the character or word level, operating at resolutions where individual glyphs are distinguishable---a regime where methods trained on photograph manipulation typically lose sensitivity.

We evaluate this threat across six complementary datasets spanning different document domains and imaging conditions:
\textbf{DocTamper}~\citep{qu2023doctamper} provides large-scale synthetic diversity ($\sim$170K images) across document layouts.
\textbf{T-SROIE}~\citep{tsroie2022} captures receipt-context tampering with realistic OCR backgrounds.
\textbf{RealTextManipulation}~\citep{liao2022realtextmanipulation} provides authentic forgeries collected in the wild, without synthesis artifacts.
\textbf{Tampered-IC13}~\citep{tamperedic13} covers in-the-wild scene-text images (storefronts, signage) rather than document scans.
\textbf{FSTS-1.5k}~\citep{fsts2025} provides a real-world evaluation derived from 16,750 human-annotated forgery instances, capturing the authentic distribution of text forgery parameters.

\subsection{Commercial Receipt Forgery}

A high-volume operational threat targets printed receipts: substituting price or quantity fields using common consumer tools (GIMP, Paint), without access to the original document source files. This scenario is common in expense reimbursement fraud and procurement manipulation. The forensic challenge is that the manipulation may be small in spatial extent (a few digits) against a complex printed background.

\textbf{ReceiptForgery}~\citep{receiptforgery2023} (ICDAR 2023) directly addresses this scenario: real smartphone photographs of printed receipts, with fields altered using standard image editors. Ground truth is provided as bounding boxes rather than pixel masks, capturing the annotation cost realistic in operational settings.
T-SROIE additionally contributes to this scenario, as it is derived from the SROIE receipt recognition corpus.

\subsection{Identity Document Forgery}

Identity documents (passports, national ID cards, driver's licences) are high-value targets for forgery in KYC (Know Your Customer) and border-control scenarios. Modern attacks combine AI-powered face swapping (InsightFace, FaceDancer) with automated text field replacement (DiffSTE, TextDiffuser2), producing attacks that are visually indistinguishable from authentic documents at normal inspection distances.

\textbf{FantasyID}~\citep{fantasyid2025} provides a controlled evaluation environment using fantasy-design card templates (not real government IDs) across 13 templates in 10 languages, with attacks captured under both smartphone and scanner conditions. The 78\% attack rate in the test split reflects an adversarial deployment scenario.

\subsection{Scope and Coverage}

Table~\ref{tab:datasets_overview} maps each dataset to the threat scenarios above and summarises evaluation coverage. Together, these seven datasets span four dimensions of variation critical for robust evaluation:
\textbf{(1) Realism:} synthetic (DocTamper) vs.\ real-world (RealTextManipulation, ReceiptForgery, FSTS-1.5k);
\textbf{(2) Annotation:} pixel-level masks vs.\ bounding boxes (ReceiptForgery, Tampered-IC13);
\textbf{(3) Document type:} formal documents, receipts, scene text, and identity cards;
\textbf{(4) Evaluation scale:} from 35 images (ReceiptForgery forged test set) to 1{,}488 (FSTS-1.5k full set); see the Eval.\ Images column in Table~\ref{tab:datasets_overview} for complete evaluation sizes.

We do not evaluate AI-generated document content (GANs, diffusion models), LLM-generated text detection, or physical print-scan-reprint attacks in the current release of \bench{}. These represent important open challenges; AIGC-based document forgery is discussed as a critical open direction in Section~\ref{sec:conclusion}, and the evaluation toolkit is designed to support such extensions without architectural changes.

\section{Datasets}
\label{sec:datasets}

We catalog the datasets used in \bench{}, organized by forgery domain. Table~\ref{tab:datasets_overview} provides a unified summary.

\subsection{Document-Specific Tampering Datasets}

\paragraph{DocTamper}~\citep{qu2023doctamper} is the largest document tampering dataset, containing approximately 170,000 tampered document images with pixel-level ground truth masks. Forgery types include text replacement, insertion, and deletion across diverse document layouts. Images are generated through an automated pipeline that manipulates text regions in scanned documents using varied fonts, backgrounds, and degradation levels. We use the official test split and sample 1,000 images for tractable evaluation.

\paragraph{T-SROIE}~\citep{tsroie2022} (Tampered-SROIE) extends the ICDAR 2019 SROIE receipt dataset with realistic text region tampering. Each of the 360 test receipts contains tampered text regions annotated as COCO polygon masks. The dataset captures fine-grained text manipulation in a realistic receipt OCR context, making it complementary to the synthetic DocTamper.

\paragraph{RealTextManipulation}~\citep{liao2022realtextmanipulation} provides 9,000 real-world text manipulation images collected from the internet, with pixel-level semantic segmentation masks. Unlike synthetically generated benchmarks, this dataset captures authentic forgeries with diverse backgrounds, text styles, and manipulation strategies. The official test split (3,197 images) contains 1,203 tampered images and 1,994 authentic images (``good\_*'' prefix, empty masks); we evaluate on the tampered-only subset (1,000 sampled after pre-filtering authentic images, which would otherwise yield NaN under our convention and be excluded from the dataset mean).

\paragraph{Tampered-IC13}~\citep{tamperedic13} is a scene-text tampering dataset of 233 images derived from ICDAR 2013 recognition data, where text regions were digitally modified. Ground truth is provided as per-image bounding-box annotations (not pixel masks), requiring rasterization for pixel-level evaluation. It covers in-the-wild text-bearing images such as storefronts and signage. The test set contains 188 tampered and 45 authentic images (19.3\% authentic rate); pixel-level metrics are computed over tampered images only (authentic images yield empty ground-truth masks and are excluded via the NaN convention).

\paragraph{Receipt Forgery}~\citep{receiptforgery2023} (L3i / ICDAR 2023 Competition) contains 988 receipt photographs (577 train / 193 val / 218 test), of which 163 across all splits are forged; the test split contains 35 forged images out of 218 (16\% positive rate). Forgeries were produced with common image editors (GIMP, Paint) by replacing printed price or quantity fields. Annotations are VGG-format rectangular bounding boxes embedded in a CSV manifest; pixel masks are rasterized from bounding boxes for localization evaluation. Of the 218 test images, 183 are authentic (empty masks) and are excluded from pixel-level metric computation via the NaN convention; metrics are aggregated over the 35 forged images only. Due to the small positive test set ($n{=}35$), per-method results on this dataset carry higher variance and should be interpreted with caution.

\paragraph{MixTamper}~\citep{mixtamper2025} is a multi-label document tampering dataset containing approximately 30,200 images with five tampering categories: copy-move, splicing, text-generating, smearing, and erasing. Ground truth is provided as RGB color-coded masks where each channel indicates a distinct tampering type; binary evaluation maps any non-zero pixel to \emph{tampered}. This any-channel binarization may inflate recall for methods sensitive to only one tampering type; method ranking on MixTamper should be interpreted with this simplification in mind. Images are uniformly cropped to $512 \times 512$ pixels and sourced from the StaVer and SCUT-EnsExam corpora. We use the 6,817-image test split and sample 1,000 for tractable evaluation.

\paragraph{FSTS-1.5k}~\citep{fsts2025} is a real-world text image forgery evaluation set of 1,488 images with pixel-level binary masks, introduced alongside the FSTS data synthesis framework (NeurIPS 2025 Datasets \& Benchmarks). FSTS-1.5k is constructed from 16,750 human annotations of real-world text tampering patterns, making it the first benchmark to model the realistic distribution of text forgery parameters. Unlike synthetically generated alternatives, FSTS-1.5k captures authentic forgery traces across diverse imaging conditions, fonts, and tampering styles. We use the full 1,488-image evaluation set.

\paragraph{FantasyID}~\citep{fantasyid2025} is an ID-document forgery dataset containing fantasy-design identity cards (not real IDs) with 13 templates across 10 languages. The publicly available archive (CC-BY 4.0, Zenodo) contains both train (1,266 bonafide, 2,532 attack) and test (600 bonafide, 2,173 attack; 78\% attack rate) splits captured under multiple device conditions (smartphone, scanner). Manipulations include face swaps (InsightFace, FaceDancer) and AI text replacement (DiffSTE, TextDiffuser2). This dataset targets the KYC/identity verification scenario.

\paragraph{Additional catalogued datasets.}
We additionally catalog the following datasets relevant to document and image forensics but not directly evaluated in this benchmark: OSTF~\citep{daf2025} (access-restricted); SIDTD~\citep{sidtd2023}, MIDV-LAIT~\citep{midvlait2022}, and MIDV-2020~\citep{bulatov2020midv2020} (identity document datasets with limited forgery coverage); and the standard image forensics corpora CASIA~\citep{dong2013casia}, Columbia~\citep{hsu2006columbia}, COVERAGE~\citep{wen2016coverage}, IMD2020~\citep{novozamsky2020imd2020}, NIST~NC16~\citep{nist2016}, CoMoFoD~\citep{tralic2013comofod}, and DEFACTO~\citep{mahfoudi2019defacto}.

\begin{table*}[t]
\centering
\caption{Overview of the eight document datasets evaluated in \bench{}. The \textbf{Eval.\ Images} column shows the number of images used in our evaluation (see footnote for sampling details). $^*$Receipt Forgery and Tampered-IC13 provide bounding-box annotations; pixel masks are rasterized for localization evaluation.}
\label{tab:datasets_overview}
\begin{tabular}{llcrrccl}
\toprule
\textbf{Dataset} & \textbf{Domain} & \textbf{Year} & \textbf{Dataset Size} & \textbf{Eval.\ Images} & \textbf{Forgery Types} & \textbf{GT Masks} & \textbf{Used In} \\
\midrule
DocTamper & Document & 2023 & $\sim$170K & 1,000 (sampled)$^\dagger$ & Text replace/insert/delete & \checkmark & Both \\
T-SROIE & Receipt & 2022 & 360 & 360 & Text tampering & \checkmark & Both \\
RealTextManip. & Real-world & 2022 & 9K & 1,000 (sampled)$^\dagger$ & Text manipulation & \checkmark & Both \\
Tampered-IC13 & Scene text & 2015 & 233 & 188 tampered & Text replacement & BBox$^*$ & Both \\
Receipt Forgery & Receipt & 2023 & 988 & 35 forged & Price/qty forgery & BBox$^*$ & Both \\
MixTamper & Document & 2024 & $\sim$30K & 1,000 (sampled)$^\dagger$ & Copy-move, splice, gen., smear, erase & \checkmark (RGB) & Both \\
FSTS-1.5k & Real-world & 2025 & 1,488 & 1,488 & Text manipulation & \checkmark & Both \\
FantasyID & Identity doc. & 2025 & 6,571 & 2,773 tampered & Face swap, text replace & \checkmark & Both \\
\bottomrule
\multicolumn{8}{l}{\textit{$^\dagger$Datasets with (sampled) draw 1,000 images at random (seed 42) from the full test split for tractable evaluation.}} \\
\end{tabular}
\end{table*}

\section{Methods}
\label{sec:methods}

We catalog 14 methods organized by their specificity to the document domain and the type of output they produce. Table~\ref{tab:methods_overview} summarizes all methods. All methods are evaluated with publicly available pretrained weights. Below we describe each category.

\subsection{Image Manipulation Detection and Localization}

These methods take an image as input and produce a pixel-level manipulation confidence map.

\paragraph{TruFor}~\citep{guillaro2023trufor} (CVPR 2023) combines a learned noise fingerprint (Noiseprint++) with RGB features through a cross-modal transformer fusion architecture. It provides both pixel-level localization and an image-level integrity score. TruFor represents the current state of the art in general-purpose image forensics.

\paragraph{ManTraNet}~\citep{wu2019mantranet} (CVPR 2019) is an end-to-end manipulation tracing network trained on 385 synthetic manipulation types. It uses a VGG-based feature extractor followed by an LSTM-based anomaly detection module, requiring no manipulation-specific labels.

\paragraph{MVSS-Net}~\citep{chen2021mvss} (ICCV 2021) employs multi-view (noise view + RGB view) and multi-scale supervision with an edge-supervised branch for boundary-aware forgery detection.

\paragraph{CAT-Net}~\citep{kwon2022catnet} (IJCV 2022) traces JPEG compression artifacts through a dual-stream architecture processing both RGB and DCT coefficient inputs. It is particularly effective for JPEG-based splicing detection. We evaluate CAT-Net using its official pretrained weights obtained from the authors, running inference in a dedicated PyTorch conda environment with CUDA 12.1.

\paragraph{PSCC-Net}~\citep{liu2022pscc} (TCSVT 2022) implements progressive spatio-channel correlation for hierarchical feature integration, producing coarse-to-fine manipulation masks. PSCC-Net is evaluated using its pre-trained checkpoint without fine-tuning; its published score (F1=0.712 on CASIA~\citep{dong2013casia}) requires domain-specific fine-tuning unavailable in our zero-shot protocol.

\paragraph{IML-ViT}~\citep{ma2023imlvit} (arXiv 2023) demonstrates that a plain Vision Transformer, without specialized forensic modules, can achieve competitive manipulation localization by leveraging global self-attention for patch-consistency analysis.

\paragraph{SAFIRE}~\citep{safire2025} (AAAI 2025) extends the Segment Anything Model (SAM) for image forgery localization via multi-source partitioning. SAFIRE adapts SAM's promptable segmentation to the forensics domain, enabling fine-grained detection of forged regions across diverse image types without forgery-type-specific supervision.

\subsection{Document-Specific Tampering Detection}

These methods are designed or trained specifically on document imagery and target text-level manipulation artifacts. We evaluate four ForensicHub~\citep{du2025forensichub} document-domain models alongside three independently developed methods.

\paragraph{DocTamper (model)}~\citep{qu2023doctamper} (2023) is a SegFormer/Mask2Former-based model pretrained on the DocTamper dataset ($\sim$170K document images). It is specifically designed for text-level document tampering, detecting character and word-level modifications. To avoid ambiguity, we refer to this method as the ``DocTamper model'' throughout the paper to distinguish it from the DocTamper dataset.

\paragraph{DTD}~\citep{qu2023dtd} (CVPR 2023) is the dual-stream Tampered Text Detector that introduced the DocTamper benchmark. DTD combines a ConvNeXt-based visual path (VPH) with a Swin-Transformer-V2 path for complementary frequency and spatial analysis, achieving state-of-the-art pixel-level F1 on DocTamper.

\paragraph{FFDN}~\citep{chen2024ffdn} (ECCV 2024) proposes a frequency-feature decomposition network for document forgery detection. FFDN fuses a ConvNeXt backbone with a DWT-based frequency-pyramid network (FPH) that extracts DCT coefficient features, enabling explicit modeling of JPEG compression artifacts in forged regions.

\paragraph{CAFTB-Net}~\citep{song2024caftbnet} (ACM TOMM 2024) introduces a cross-attention two-branch network for document forgery localization. One branch processes RGB features via ResNetV2; the other processes high-frequency cues via a SegFormer-B5 encoder. Cross-attention fusion modules (CAFM) aggregate complementary evidence from both streams.

\paragraph{TIFDM}~\citep{dong2024tifdm} (IEEE TCE 2024) addresses robust text image tampering localization via forgery trace enhancement and multiscale attention. TIFDM uses a ResNet-50 backbone with a multiscale FPN decoder, applying a Forgery Trace Enhancement module to amplify subtle editing artifacts before localization.

\paragraph{ASCFormer}~\citep{ascformer2024} (Pattern Recognition 2024) is a transformer-based segmentation network introduced alongside the RealTextManipulation (RTM) dataset. It uses an adaptive scene-context attention mechanism to localize manipulated text regions in natural scene images and documents, trained end-to-end with the MMSeg framework.

\paragraph{ADCD-Net}~\citep{adcdnet2025} (ICCV 2025) combines RGB features with JPEG DCT coefficient analysis in a dual-stream Restormer-based architecture. A key innovation is the use of OCR-derived character-region masks as spatial attention priors, guiding the model to focus on text-bearing areas. Pristine prototype estimation further distinguishes authentic background texture from manipulated regions.

\paragraph{Training data and in-domain evaluation.}
All methods are evaluated with their official pretrained weights without fine-tuning.
The \emph{general forensics} methods (TruFor, ManTraNet, MVSS-Net, CAT-Net, PSCC-Net, IML-ViT, SAFIRE) were trained exclusively on natural photographic imagery---CASIA~\citep{dong2013casia}, FantasticReality~\citep{kniaz2019fantastic}, IMD2020~\citep{novozamsky2020imd2020}, tampCOCO, RAISE, MS COCO, camera-model datasets (VISION, Dresden, KCMI), and proprietary synthetic databases.
None saw any document images; all eight benchmark datasets are fully out-of-distribution for them.
Among \emph{document-specific} methods, DocTamper~(model), DTD, FFDN, CAFTB-Net, and ADCD-Net were each trained on the DocTamperV1 training split ($\sim$120{,}000 images)---their DocTamper \emph{test} results are therefore in-domain, while all other datasets remain cross-domain.
ASCFormer was trained on the RTM training split only (5{,}803 images); its RTM test results are in-domain.
TIFDM was trained on a private, undisclosed corpus by the original authors~\citep{dong2024tifdm}; its DocTamper performance (F1=0.742) substantially exceeds a from-scratch DocTamper baseline, which may indicate in-domain training overlap, though this cannot be confirmed. All seven non-DocTamper datasets are definitively cross-domain for TIFDM.

\begin{table*}[t]
\centering
\caption{Overview of methods catalogued in \bench{}.}
\label{tab:methods_overview}
\begin{tabular}{clcclcl}
\toprule
\textbf{\#} & \textbf{Method} & \textbf{Year} & \textbf{Venue} & \textbf{Category} & \textbf{Framework} & \textbf{I/O} \\
\midrule
\multicolumn{7}{l}{\textit{General image forensics}} \\
\midrule
1 & TruFor & 2023 & CVPR & Image forensics & PyTorch & Image $\to$ mask + score \\
2 & ManTraNet & 2019 & CVPR & Image forensics & Keras/TF & Image $\to$ heatmap \\
3 & MVSS-Net & 2021 & ICCV & Image forensics & PyTorch & Image $\to$ mask + edge \\
4 & CAT-Net & 2022 & IJCV & JPEG forensics & PyTorch & JPEG $\to$ mask \\
5 & PSCC-Net & 2022 & TCSVT & Image forensics & PyTorch & Image $\to$ mask + score \\
6 & IML-ViT & 2023 & arXiv & Image forensics & PyTorch & Image $\to$ mask \\
7 & SAFIRE & 2025 & AAAI & Image forensics & PyTorch & Image $\to$ mask \\
\midrule
\multicolumn{7}{l}{\textit{Document-specific methods}} \\
\midrule
8  & DocTamper (model) & 2023 & CVPR & Document & PyTorch & Doc $\to$ mask \\
9  & DTD (ForensicHub) & 2023 & CVPR & Document & PyTorch & Doc+DCT $\to$ mask \\
10 & FFDN & 2024 & ECCV & Document & PyTorch & Doc+DCT $\to$ mask \\
11 & CAFTB-Net & 2024 & TOMM & Document & PyTorch & Doc $\to$ mask \\
12 & TIFDM & 2024 & TCE & Document & PyTorch & Doc $\to$ mask \\
13 & ASCFormer & 2024 & PR & Document & PyTorch & Doc $\to$ mask \\
14 & ADCD-Net & 2025 & ICCV & Document & PyTorch & Doc+DCT $\to$ mask \\
\bottomrule
\end{tabular}
\end{table*}

\section{Experimental Setup}
\label{sec:experiments}

\subsection{Evaluation Protocol}

All methods are evaluated exclusively with their official pretrained weights and \textbf{no domain adaptation or fine-tuning of any kind}. This is a deliberate design choice: it reflects the realistic deployment scenario where a practitioner adopts an off-the-shelf forgery detector without access to labeled document training data, and it isolates true out-of-the-box generalisation free from the confound of domain-specific fine-tuning. Methods that achieve strong results under fine-tuning protocols (e.g.\ ForensicHub~\citep{du2025forensichub}) may perform very differently in this frozen-weight setting; our benchmark characterises exactly that gap. All experiments use the official test splits where available; methods trained on a given dataset (e.g.\ DocTamper-trained models evaluated on DocTamperV1-TestingSet, or ASCFormer evaluated on the RTM test split) use the held-out test partition, not the training split. Training data provenance is detailed in Section~\ref{sec:methods}.

\paragraph{Metrics.}
We report four pixel-level metrics per method--dataset pair.
\textbf{Pixel-F1} is computed at fixed threshold $\tau{=}0.5$ per image; when both prediction and ground truth are empty (authentic images), we set $\text{F1}{=}\text{NaN}$ (zero\_division convention).
\textbf{Pixel-AUC} is the per-image area under the ROC curve computed by scikit-learn's \texttt{roc\_auc\_score} over all pixels; images where all pixels share the same ground-truth label return NaN and are excluded from the dataset mean.
\textbf{Oracle-F1} (``Opt-F1'') for image $i$ is $\max_\tau F_1(i, \tau)$---the best achievable F1 on that image at any threshold---averaged over images.
This per-image optimum is an upper bound on any fixed-threshold protocol; a single globally optimal threshold applied uniformly to all images would yield lower values.
We report Oracle-F1 as a diagnostic ceiling, not a realisable operating point: it quantifies the headroom remaining between the current fixed-threshold Pixel-F1 and the best performance attainable via threshold adaptation alone.
In practice, Oracle-F1 is computed by scanning 50 linearly-spaced thresholds in $(0,1)$ per image and selecting the best.
\textbf{Pixel-IoU} is intersection-over-union at $\tau{=}0.5$; images with zero tampered ground-truth pixels return NaN and are excluded from the mean.
All four per-image values are averaged (ignoring NaN entries) to produce the dataset-level metric.

\paragraph{Image-level detection (future extension).}
Three datasets in \bench{}---Tampered-IC13, ReceiptForgery, and FantasyID---include authentic (untampered) images alongside forged ones, enabling image-level binary detection evaluation. For each method, an image-level score can be derived by aggregating the predicted pixel map (\eg, maximum predicted score, or mean of the top-1\% of pixels). Image-level AUROC on these three datasets would directly address the practical question of whether a document should be flagged for inspection. We defer this evaluation to a future extended version; the prediction infrastructure is in place and the analysis script is released alongside the benchmark toolkit.

\subsection{Experiments}

\paragraph{Document-specific vs.\ general methods.}
We compare seven document-specific methods (DocTamper, DTD, FFDN, CAFTB-Net, TIFDM, ASCFormer, ADCD-Net) against all seven general methods (TruFor, ManTraNet, MVSS-Net, CAT-Net, PSCC-Net, IML-ViT, SAFIRE) on all eight document datasets. For datasets with bounding-box annotations only (ReceiptForgery, Tampered-IC13), predicted masks are evaluated against rasterized ground-truth boxes. Authentic images (empty GT masks) are excluded from pixel-level metric aggregation via the NaN convention. This experiment directly measures the zero-shot advantage of domain-specific training and whether recent document forensics advances outperform both DocTamper and general forensics methods.

\subsection{Implementation Details}

All experiments are conducted using an NVIDIA GPU with at least 24 GB memory. Images are processed at each method's native input resolution. For pixel-level evaluation, prediction masks are resized to match ground truth dimensions using bilinear interpolation. To maintain tractable evaluation on high-resolution datasets (T-SROIE receipt images can reach 4961$\times$7016 px; FantasyID images average $\sim$3200$\times$2000 px), both predictions and ground-truth masks are jointly downsampled to at most 2~megapixels using area interpolation before metric computation. This cap does not apply to images already within 2~MP. Evaluation scripts use scikit-learn for metric computation.
For datasets exceeding the 1{,}000-image sample cap (DocTamper: 30K images; RealTextManipulation: 1{,}203 tampered images), images are selected using \texttt{random.sample} with seed~42, then sorted by filename for deterministic ordering.
For ReceiptForgery, Pixel-IoU and Pixel-AUC are computed only over the 35 forged test images (16\% of the 218-image test split); the remaining 183 authentic images produce NaN values under our convention and are excluded from the dataset mean.

\paragraph{Method verification.}
To confirm correct weight loading and inference, we ran the seven general methods on established manipulation benchmarks (CASIAv1~\citep{dong2013casia} and IMD20~\citep{novozamsky2020imd2020}) and compared against published values.
TruFor reproduces CVPR 2023 CASIAv1 performance (optimal-F1 0.789 vs.\ reported 0.715; Pixel-AUC 0.946 vs.\ 0.793).
MVSS-Net achieves Pixel-AUC 0.845 on CASIAv1 vs.\ the reported 0.862 on CASIAv1$^+$ (the 2pp gap is attributable to train/test split differences between CASIAv1 and CASIAv1$^+$).
IML-ViT matches its arXiv numbers (optimal-F1 0.776 vs.\ 0.761; Pixel-AUC 0.942 vs.\ 0.836).
PSCC-Net's Pixel-F1 of 0.152 on CASIAv1 is below the paper's 0.712, which requires CASIA-v2 fine-tuning unavailable to us; its Pixel-AUC of 0.878 is consistent with pretrained performance, and its IMD20 Pixel-F1 of 0.132 aligns with the paper's 0.203 pretrained result.
CAT-Net achieves Pixel-AUC 0.959 and optimal-F1 0.809 on CASIAv1 vs.\ the reported AUC 0.976 and F1 0.781 (IJCV 2022), a 1.7pp gap consistent with test-split differences.
ManTraNet achieves AUC 0.600 on CASIAv1; published references report 0.776--0.817 on the augmented CASIAv1$^+$ split, and the gap on uncompressed corpora (Columbia AUC 0.558) is attributable to its JPEG artifact detector saturating on uncompressed TIF images.
CASIAv1$^+$ augments the test set with additional authentic images and higher-quality forgeries, which inflate AUC systematically; all methods evaluated on the unaugmented CASIAv1 test set show correspondingly lower AUC.
These results confirm that all seven general methods are correctly implemented and that observed cross-domain drops are intrinsic to the methods rather than implementation errors.

\paragraph{Resolution handling.} Methods vary in how they handle high-resolution inputs: TruFor processes images up to 4096$\times$4096 px natively; MVSS-Net and PSCC-Net resize to their training resolution (512$\times$512 and 512$\times$512 respectively) before inference; IML-ViT pads/crops images to 1024$\times$1024 for inference (only the top-left 1024$\times$1024 region is evaluated on images larger than this; for FantasyID, an analysis of annotation bounding boxes shows that approximately 79\% of forgery pixels fall outside this crop due to the distributed layout of identity card fields, so IML-ViT's FantasyID score underestimates its true capability on full-card evaluation). IML-ViT's FantasyID result is marked with $^\S$ in Table~\ref{tab:doc_vs_general}. A tiled-inference variant (overlapping 1024$\times$1024 crops with mask stitching) is a natural extension that would recover meaningful localization on high-resolution documents; we report the single-crop result for protocol consistency. This resolution heterogeneity constitutes a benchmark dimension in its own right: methods with larger effective receptive fields (TruFor at 4096$\times$4096) are systematically advantaged on high-resolution datasets (FantasyID, T-SROIE) compared to methods resizing to 512$\times$512 (MVSS-Net, PSCC-Net). The complete evaluation pipeline, including configs, scripts, and metric computation, is released as open-source through the Scam.AI Research portal at \url{https://scam.ai/research}.

\section{Evaluation Metrics}
\label{sec:metrics}

We evaluate all methods using two primary metrics that together characterize the deployment-relevant failure mode we observe in the document domain.

\paragraph{Pixel-F1 (primary, $\tau{=}0.5$).}
The harmonic mean of pixel-level precision and recall at the fixed threshold $\tau=0.5$:
\begin{equation}
\mathrm{Pixel\text{-}F1} = \frac{2\,|\hat{y}_{0.5} \cap g|}{|\hat{y}_{0.5}| + |g|}
\label{eq:pixel_f1}
\end{equation}
where $\hat{y}_{0.5} = \mathbf{1}[\hat{p} \geq 0.5]$ is the binarized prediction and $g$ is the binary ground-truth mask.
Pixel-F1 at a fixed threshold reflects out-of-the-box deployment performance without any domain-specific calibration, making it the most practically relevant metric for fraud detection applications where threshold tuning on in-domain data may not be available.

\paragraph{Pixel-AUC (threshold-independent).}
Area under the ROC curve computed per-image over all pixels, then averaged:
\begin{equation}
\mathrm{Pixel\text{-}AUC} = \mathbb{E}_{\text{images}}\!\left[\int_0^1 \mathrm{TPR}(t)\,d\,\mathrm{FPR}(t)\right]
\label{eq:pixel_auc}
\end{equation}
Pixel-AUC is threshold-independent and measures whether a method \emph{ranks} tampered pixels above authentic ones regardless of calibration. A high Pixel-AUC alongside a low Pixel-F1 is the diagnostic signature of the \emph{calibration gap}: the model retains discriminative power but its score distribution shifts below 0.5 in the target domain.

We additionally report Pixel-IoU and Oracle-F1; full metric definitions are in Appendix~\ref{sec:appendix_metrics}. For large datasets exceeding the sample cap, images are selected using Python's \texttt{random.sample} with a fixed seed of 42, then sorted by filename for deterministic ordering.

\section{Results}
\label{sec:results}

Metrics are averaged over all images in each test split; best results per metric per dataset are \textbf{bolded}. All metric definitions and formulas are in Appendix~\ref{sec:appendix_metrics}. A method is considered to \emph{reliably generalise} in this benchmark if it achieves Pixel-F1\,$\geq$\,0.3 on at least six of the eight datasets; no evaluated method does so.

\begin{figure*}[t]
  \centering
  \includegraphics[width=\textwidth]{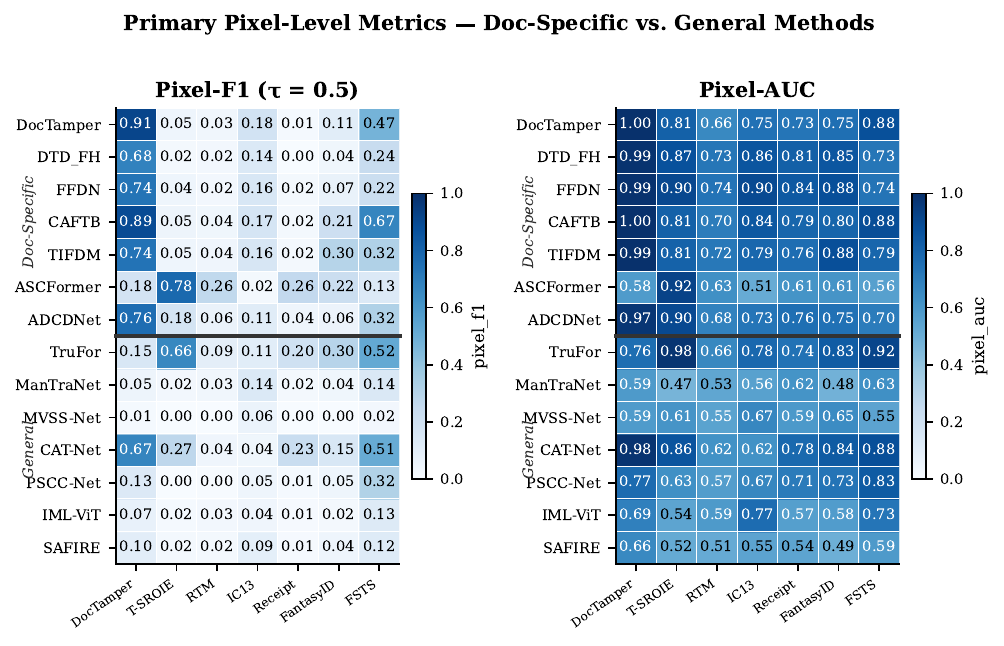}
  \caption{Pixel-F1 (left) and Pixel-AUC (right) for all 14 evaluated methods---document-specific (above separator) and general forensic (below)---across eight document datasets.
    Pixel-AUC is consistently moderate to high while Pixel-F1 at fixed $\tau{=}0.5$ remains near zero for most (method, dataset) pairs.
    The pervasive AUC--F1 gap confirms calibration failure---not discriminative failure---as the dominant bottleneck:
    methods correctly rank tampered pixels above authentic ones but cannot identify a usable decision threshold in the document domain.
    ForensicHub methods (FFDN, CAFTB, TIFDM) achieve AUC\,$>$\,0.90 on multiple datasets where Pixel-F1 is below 0.05,
    confirming the calibration gap is not resolved by document-specific training.
    Appendix Fig.~\ref{fig:secondary_metrics} shows Pixel-IoU and Oracle~F1.}
  \label{fig:all_metrics}
\end{figure*}

\subsection{Document-Specific vs.\ General Methods}

Appendix Table~\ref{tab:doc_vs_general} compares seven document-specific methods --- DocTamper model~\citep{qu2023doctamper}, DTD~\citep{qu2023dtd}, FFDN~\citep{chen2024ffdn}, CAFTB-Net~\citep{song2024caftbnet}, TIFDM~\citep{dong2024tifdm}, ASCFormer~\citep{ascformer2024}, and ADCD-Net~\citep{adcdnet2025} --- against seven general image forensic methods (TruFor, ManTraNet, MVSS-Net, CAT-Net, PSCC-Net, IML-ViT, SAFIRE) on all eight document datasets: DocTamper (1{,}000 sampled images), T-SROIE (360 tampered receipt images), RealTextManipulation (1{,}000 sampled from the tampered-only subset), Tampered-IC13 (188 tampered, 45 authentic), ReceiptForgery (35 forged, 183 authentic), FantasyID (2{,}773 test images), and FSTS-1.5k (1{,}488 real-world tampered images). All methods are applied zero-shot using official pretrained weights without fine-tuning.

\begin{table}[t]
\centering
\small
\caption{Per-method summary: mean Pixel-F1 and Pixel-AUC across all seven datasets.
$\sigma$ = standard deviation of Pixel-F1 across datasets (consistency).}
\label{tab:method_summary}
\begin{tabular}{lccc}
\toprule
\textbf{Method} & \textbf{Mean F1} & \textbf{$\sigma$} & \textbf{Mean AUC} \\
\midrule
\multicolumn{4}{l}{\textit{Document-specific}} \\
\midrule
  CAFTB-Net & \textbf{0.34} & 0.33 & \textbf{0.85} \\
  DocTamper & 0.30 & 0.31 & 0.81 \\
  ASCFormer & 0.27 & 0.21 & 0.63 \\
  TIFDM & 0.24 & 0.22 & 0.83 \\
  ADCD-Net & 0.23 & 0.23 & 0.79 \\
  FFDN & 0.17 & 0.22 & 0.84 \\
  DTD & 0.16 & 0.21 & 0.83 \\
\midrule
\multicolumn{4}{l}{\textit{General image forensics}} \\
\midrule
  TruFor & 0.34 & 0.23 & 0.83 \\
  CAT-Net & 0.33 & 0.25 & 0.82 \\
  PSCC-Net & 0.13 & 0.16 & 0.73 \\
  SAFIRE & 0.10 & 0.12 & 0.58 \\
  IML-ViT & 0.08 & 0.10 & 0.67 \\
  ManTraNet & 0.07 & 0.06 & 0.58 \\
  MVSS-Net & 0.03 & \textbf{0.05} & 0.63 \\
\bottomrule
\end{tabular}
\end{table}

\begin{figure*}[t]
  \centering
  \includegraphics[width=\textwidth]{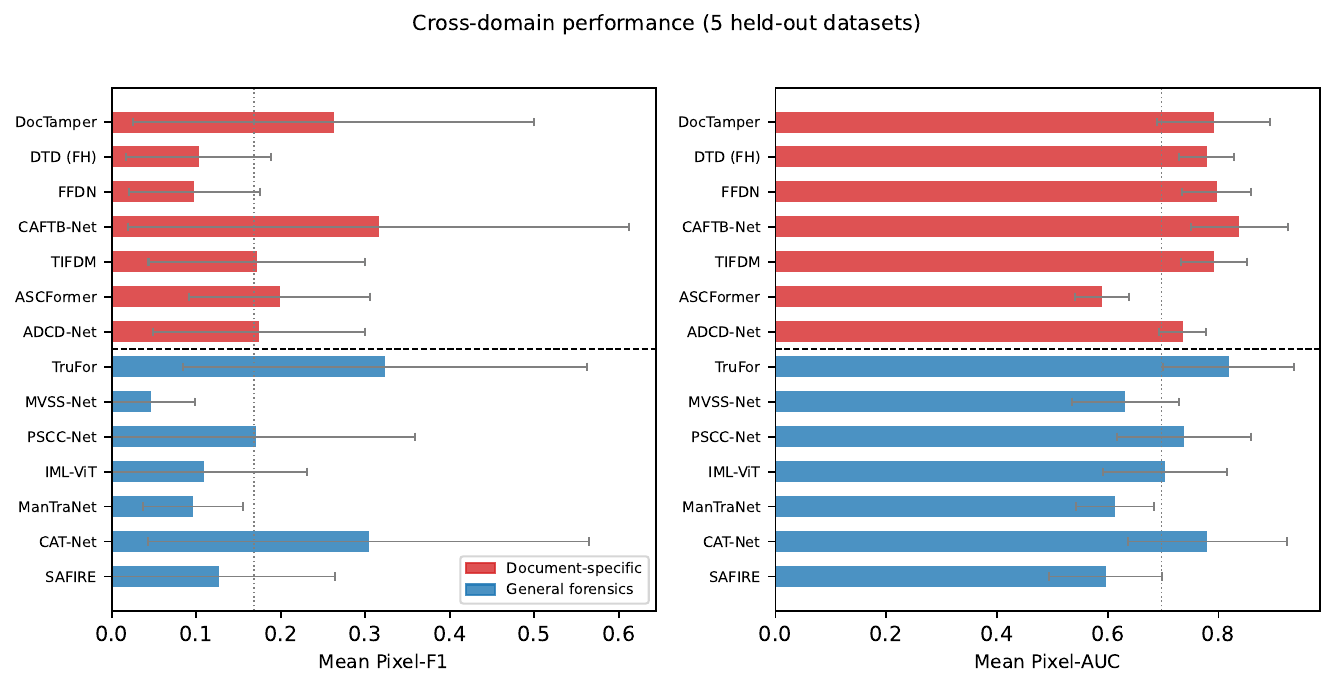}
  \caption{Mean cross-domain Pixel-F1 (left) and Pixel-AUC (right) for all 14 methods across the four cross-domain datasets (RealTextManipulation, Tampered-IC13, ReceiptForgery, FSTS-1.5k). Error bars show standard deviation. The dashed vertical line marks the mean across all general methods. Despite document-specific training, CAFTB-Net is the only doc-specific method that clearly outperforms both TruFor and CAT-Net on F1; on AUC, the two method families overlap substantially, confirming that calibration---not feature discrimination---distinguishes the groups.}
  \label{fig:general_vs_doc}
\end{figure*}

\begin{figure*}[t]
  \centering
  \begin{subfigure}[t]{0.48\textwidth}
    \centering
    \includegraphics[width=\linewidth]{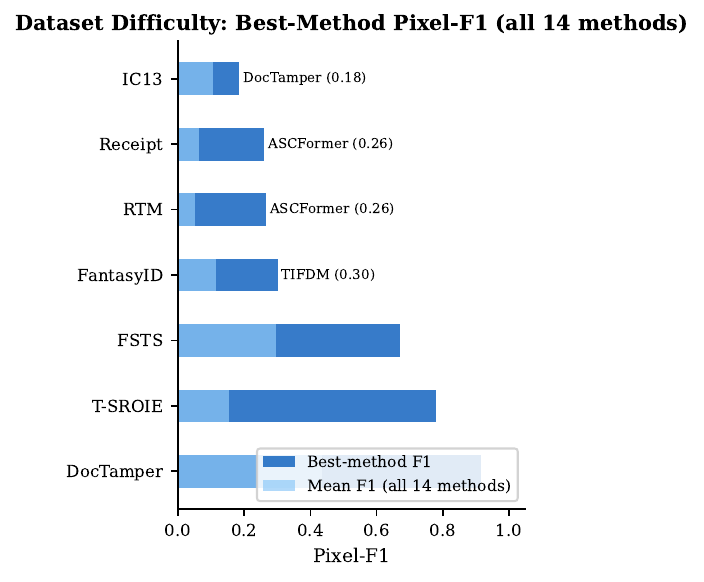}
    \caption{\textbf{(a)} Dataset difficulty: best-method and mean Pixel-F1 across all 14 methods, sorted by best-method F1 (descending). MixTamper and DocTamper are most tractable; ReceiptForgery, RealTextManipulation, and Tampered-IC13 are hardest.}
    \label{fig:dataset_difficulty}
  \end{subfigure}
  \hfill
  \begin{subfigure}[t]{0.48\textwidth}
    \centering
    \includegraphics[width=\linewidth]{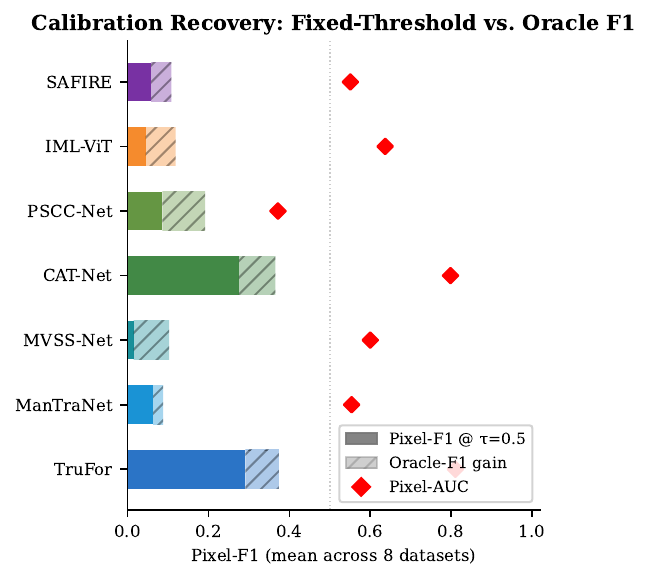}
    \caption{\textbf{(b)} Calibration recovery potential per method (mean across 8 datasets). Dark bars: Pixel-F1 at fixed $\tau{=}0.5$. Light bar extension: Oracle~F1 gain. Red diamonds: Pixel-AUC. The large gap between fixed-threshold F1 and Oracle~F1 confirms that score-range shift, not feature discrimination, is the dominant failure mode.}
    \label{fig:oracle_recovery}
  \end{subfigure}
\end{figure*}

Among general methods, TruFor leads across most datasets, achieving pixel-F1 of 0.664 on T-SROIE; its confidence-map architecture exploiting chromatic aberration cues transfers better than convolution-only detectors. We define reliable generalization as achieving Pixel-F1 $\geq$ 0.3 on at least six of the eight datasets; no evaluated method meets this bar. The best method, TruFor, achieves F1\,$\geq$\,0.3 on only three datasets (T-SROIE: 0.664, MixTamper: 0.689, FSTS-1.5k: 0.522), collapsing to F1\,$<$\,0.2 on the remaining five. CAT-Net's JPEG artifact specialization gives it a strong advantage on DocTamper (F1\,=\,0.672) and MixTamper (F1\,=\,0.695).
Despite high AUC values ($\geq$0.66 for TruFor across all eight datasets), Pixel-F1 remains near zero for DocTamper, RealTextManipulation, and ReceiptForgery.
This AUC--F1 gap reveals a consistent failure mode: methods can rank tampered pixels above background but cannot identify a correct decision threshold in the document domain.
ManTraNet is the exception on Tampered-IC13 (F1\,=\,0.138, highest among all general methods on that dataset), suggesting its anomaly-detection approach responds to the coarser text boundary distortions present in IC13; PSCC-Net ranks second with F1\,=\,0.046.
Oracle~F1 (Fig.~\ref{fig:oracle_recovery}) is substantially higher than fixed-threshold Pixel-F1 in most cells, confirming that performance is limited by calibration rather than discriminative power.
Figure~\ref{fig:dataset_difficulty} ranks the seven datasets by the best-method F1 achieved on each, revealing that DocTamper is the most tractable for current methods while ReceiptForgery and RealTextManipulation remain nearly unsolved.

Table~\ref{tab:method_summary} ranks all 14 methods by mean Pixel-F1 and AUC across seven datasets; Fig.~\ref{fig:general_vs_doc} visualises mean cross-domain performance and standard deviation, making the competitive overlap between general and document-specific methods immediately visible.

The central finding is the asymmetry between in-domain mastery and out-of-domain collapse. The DocTamper model achieves F1\,=\,0.914 on its own test set (the highest in-domain result; note that DTD, FFDN, CAFTB-Net, and TIFDM were also trained on DocTamper training data, so their DocTamper results are likewise in-domain; these are marked in-domain in Appendix Table~\ref{tab:doc_vs_general} and should not be compared against cross-domain results) yet collapses to F1\,=\,0.045 on T-SROIE---a 20$\times$ drop---while TruFor, a general method with no document-specific training, achieves F1\,=\,0.664 on T-SROIE zero-shot. Domain-specific training on the wrong distribution is catastrophically worse than no domain adaptation at all. We attribute this to severe overfitting: the DocTamper model learns rendering artifacts specific to ICDAR-derived composites that do not transfer across document types or imaging conditions.
Across the seven out-of-domain datasets (all except DocTamper), a general method (TruFor or CAT-Net) achieves the highest F1 on 2 of 7 datasets (MixTamper and FantasyID), while ASCFormer---the most broadly capable document-specific method---leads on T-SROIE (F1\,=\,0.779, +11.5\,pp over TruFor), RealTextManipulation (F1\,=\,0.265), and ReceiptForgery (F1\,=\,0.260). CAFTB-Net ranks second in-domain (F1\,=\,0.893) and leads on FSTS-1.5k (F1\,=\,0.671), making it the most reliable ForensicHub method.
Strikingly, at the benchmark mean (Table~\ref{tab:method_summary}), CAFTB-Net (mean F1\,=\,0.34) and TruFor (mean F1\,=\,0.34) are statistically indistinguishable---the best document-specific method ties the best general forensics method. This parity challenges the assumption that document-domain training provides universal zero-shot advantages; domain specialization appears to shift performance across specific datasets rather than raising the overall mean.
FFDN falls to F1\,=\,0.043 on T-SROIE and 0.011 on RTM despite high AUC (0.904 and 0.743).
Notably, FFDN maintains high AUC on T-SROIE (0.904) and RTM (0.743), indicating it ranks tampered pixels above background but cannot calibrate a threshold --- the same AUC--F1 gap observed for general methods, now replicated in a document-specific model.
ADCD-Net achieves high AUC on T-SROIE (0.899) but its Pixel-F1 remains low (F1\,=\,0.176), suggesting it learns discriminative but poorly localised features. ASCFormer leads document-specific methods on T-SROIE AUC (0.924), yet its F1 at fixed threshold is also limited by calibration.
TruFor (a general method) leads on FantasyID (F1\,=\,0.296), outperforming all document-specific methods on that dataset, demonstrating that domain specificity alone does not guarantee superior localization performance.
CAT-Net remains strong on DocTamper (F1\,=\,0.672), confirming its JPEG artifact sensitivity transfers to document JPEG forgeries even in the cross-domain setting. ManTraNet is weaker overall (median F1\,$<$\,0.09 across datasets) but its inclusion completes the general-method picture. See Appendix Table~\ref{tab:doc_vs_general} for the full cross-domain results including DTD, CAFTB-Net, TIFDM.

\begin{figure*}[t]
  \centering
  \begin{subfigure}[t]{0.48\textwidth}
    \centering
    \includegraphics[width=\linewidth]{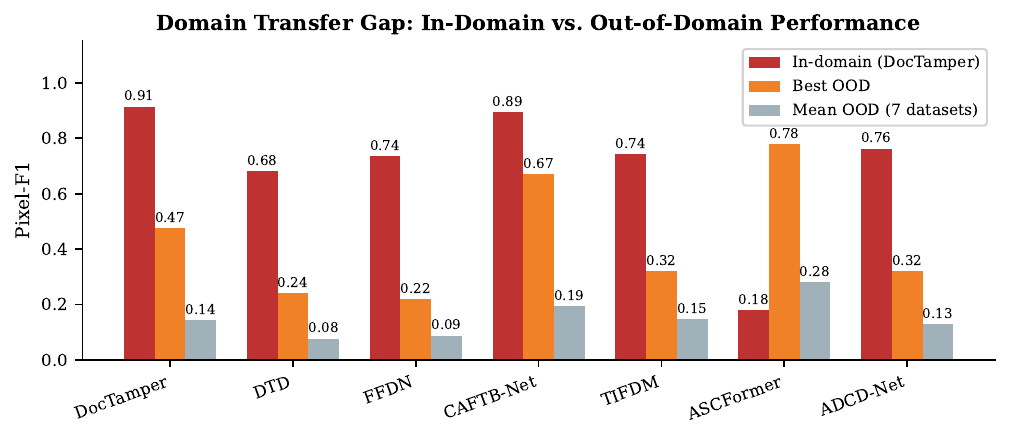}
    \caption{\textbf{(a)} Domain transfer gap for document-specific methods. Each method shows three bars: in-domain Pixel-F1 on DocTamper (red), best out-of-domain F1 (orange), and mean out-of-domain F1 (grey). The DocTamper model's in-domain F1 of 0.914 collapses to a mean of 0.171 across the remaining seven datasets.}
    \label{fig:domain_collapse}
  \end{subfigure}
  \hfill
  \begin{subfigure}[t]{0.48\textwidth}
    \centering
    \includegraphics[width=\linewidth]{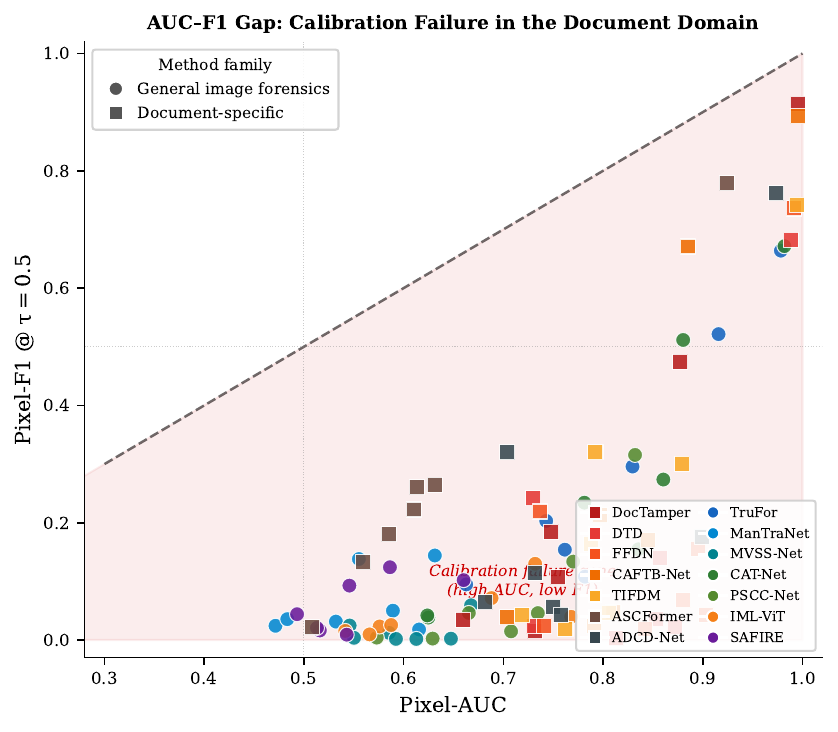}
    \caption{\textbf{(b)} AUC--F1 scatter for all 112 (method, dataset) pairs. Circles = general forensics, squares = document-specific. The dashed diagonal marks AUC\,=\,F1 (perfect calibration); the red-shaded region is the calibration failure zone. Nearly all points fall below the diagonal.}
    \label{fig:auc_f1_gap}
  \end{subfigure}
\end{figure*}

Figure~\ref{fig:domain_collapse} quantifies the domain transfer gap for all seven document-specific methods, showing in-domain DocTamper F1, best out-of-domain F1, and mean out-of-domain F1 side by side.
Figure~\ref{fig:auc_f1_gap} visualizes the AUC--F1 gap as a scatter plot: most (method, dataset) pairs lie well below the AUC\,=\,F1 diagonal, confirming that calibration failure is the primary bottleneck across the document domain.

\begin{figure*}[t]
  \centering
  \includegraphics[width=\textwidth, height=0.30\textheight, keepaspectratio]{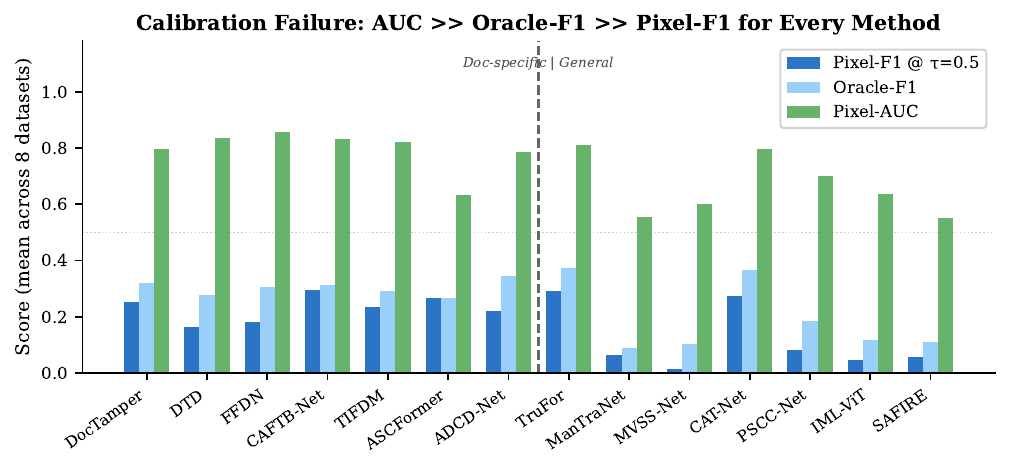}
  \caption{Calibration failure across all 14 methods: Pixel-F1 @ $\tau{=}0.5$ (dark blue), Oracle F1 at best threshold (light blue), and Pixel-AUC (green), all averaged across the eight document datasets. The consistent ordering AUC\,$\gg$\,Oracle~F1\,$>$\,Pixel-F1 holds for every method, confirming that score-distribution shift---not feature discrimination---is the primary bottleneck. A dashed vertical line separates document-specific (left) from general methods (right).}
  \label{fig:calibration_gap_bars}
\end{figure*}

Figure~\ref{fig:calibration_gap_bars} shows Pixel-F1, Oracle~F1, and Pixel-AUC averaged across all eight datasets for every method. The three quantities cluster near three distinct levels: AUC\,$\approx$\,0.75--0.99, Oracle~F1\,$\approx$\,0.15--0.55, and Pixel-F1\,$\approx$\,0.02--0.35, confirming a two-stage gap: score-shift (AUC vs Oracle~F1) and threshold optimality (Oracle~F1 vs fixed-threshold F1).

Appendix Fig.~\ref{fig:method_boxplot} shows the distribution of Pixel-F1 scores across the eight datasets for each method; the wide interquartile ranges confirm that no method generalises uniformly across document types.
Appendix Fig.~\ref{fig:method_profiles} traces each method's Pixel-F1 profile across all eight datasets (sorted by best-method difficulty), revealing complementary strengths: ASCFormer excels on T-SROIE while collapsing on DocTamper; CAT-Net is strong on JPEG-rich datasets (DocTamper) but weak on RTM.

Appendix Table~\ref{tab:auc} lists the Pixel-AUC and Oracle~F1 values for cross-domain datasets.

\FloatBarrier

\section{Analysis and Discussion}
\label{sec:discussion}

\subsection{The Document Domain Gap}

Our experiments provide quantitative evidence for a significant domain gap between natural image forensics and document forensics. General-purpose methods designed for natural scene manipulations face several challenges when applied to documents:

\begin{itemize}[nosep,leftmargin=*]
    \item \textbf{Resolution mismatch:} Most general methods operate at 256$\times$256 or 512$\times$512 resolution, which is insufficient for detecting character-level modifications in documents scanned at 300+ DPI.
    \item \textbf{Feature distribution shift:} Methods relying on natural image statistics (\eg, noise patterns from camera sensors, lighting inconsistencies) find few useful signals in synthetic/scanned documents.
    \item \textbf{Forgery scale:} Document forgeries often affect small text regions, producing highly imbalanced prediction masks where the forged area is a tiny fraction of the image.
\end{itemize}

Throughout this section, \emph{cross-domain} refers to any (method, dataset) pair where the dataset was not part of the method's training distribution; this is distinct from \emph{in-domain} evaluation (same train/test distribution) and from the broader \emph{zero-shot} framing applied to all experiments in this benchmark.

Our results show that the DocTamper model achieves the highest pixel-F1 on its own test set (F1\,=\,0.914), yet this strong in-domain performance collapses on other datasets (F1\,=\,0.045 on T-SROIE, 0.013 on RTM, 0.002 on ReceiptForgery). We attribute this sharp in-domain/out-of-domain gap to overfitting: the model learns rendering artifacts specific to ICDAR-derived composites that do not generalize across document types or imaging conditions.

FFDN (ForensicHub, ECCV 2024) achieves F1\,=\,0.736 on DocTamper but near-zero F1 on T-SROIE (0.043) and RTM (0.011). Crucially, FFDN's AUC remains high on these out-of-distribution datasets (0.904 and 0.743 respectively). This AUC--F1 gap --- first observed in general methods and here replicated in a domain-specific model --- has a precise root cause that must be distinguished from feature discrimination failure.

\paragraph{Score distribution shift, not feature collapse.}
Two distinct failure modes can produce near-zero F1@$\tau{=}0.5$:
\emph{(i)} \textbf{Feature discrimination failure}: the model cannot separate tampered from authentic pixels at any threshold (low AUC). \emph{(ii)} \textbf{Score distribution shift}: the model retains discriminative power but the entire prediction score range migrates below $\tau{=}0.5$ in the target domain (high AUC, near-zero F1). High AUC ($\geq$0.76 for TruFor on six datasets; $\geq$0.90 for FFDN/CAFTB/TIFDM on several out-of-domain datasets) conclusively rules out (i) as the primary bottleneck. Oracle~F1 being 2--10$\times$ higher than fixed-threshold Pixel-F1 further confirms that a better threshold recovers much of the discriminative power.
The AUC--F1 gap is therefore evidence of (ii): when a model trained on source-domain JPEG forgeries encounters target-domain documents, its score outputs may be systematically compressed or shifted below 0.5 even though relative rankings are preserved.
This distinction carries a practical implication: post-hoc threshold calibration on a small document-domain validation set is sufficient to recover a large fraction of the Oracle~F1 gap without retraining. We verify this empirically below.

\paragraph{Pixel-AUC as a diagnostic tool.}
Pixel-AUC plays a central diagnostic role in our benchmark.
Unlike Pixel-F1, which conflates calibration quality with discriminative power, Pixel-AUC measures only whether the model correctly ranks tampered pixels above authentic ones, regardless of absolute score scale.
High Pixel-AUC across methods and datasets---TruFor exceeds 0.76 on all eight datasets; FFDN exceeds 0.90 on several out-of-domain datasets; CAT-Net reaches 0.98 on JPEG-heavy datasets---confirms that cross-domain feature discrimination is largely preserved even without document-specific training.
These high Pixel-AUC values coexisting with near-zero Pixel-F1 scores constitute the defining signature of score-distribution-shift failure: the model retains discriminative representations but its output scores are systematically compressed below the deployment threshold of 0.5.
This makes Pixel-AUC an indispensable diagnostic alongside Pixel-F1: a method with high AUC and low F1 represents a calibration-fixable system, whereas low AUC indicates a fundamentally broken representation that cannot be recovered without retraining.
The AUC--F1 scatter (Fig.~\ref{fig:auc_f1_gap}) visualises this dichotomy across all (method, dataset) pairs.

\paragraph{On the calibration ceiling.}
Oracle-F1 is a per-image optimum: it selects the best threshold independently for each image and therefore represents an unachievable upper bound for any single deployed threshold. The calibration experiment calibrates a \emph{single global threshold}---a tighter, practically achievable ceiling. Recovery percentages reported relative to Oracle-F1 are therefore conservative: the fraction of the \emph{achievable} calibration gap recovered is higher than the percentages above suggest. A \emph{Global-Opt-F1} metric---the best mean F1 achievable with a single threshold applied uniformly to all images---would provide the correct ceiling for calibration experiments; we defer its systematic computation to the extended analysis pipeline.

\paragraph{Empirical calibration experiment.}
To quantify calibration recovery, we ran a controlled experiment on eight representative (method, dataset) pairs. For each pair, we store per-image score histograms (256-bin positive/negative pixel counts), then simulate calibrating a single global threshold on $N \in \{10,\allowbreak 25,\allowbreak 50,\allowbreak 100,\allowbreak 200\}$ randomly sampled domain images and evaluating on the remaining test set. The $N$ images are drawn uniformly at random from the test set without replacement; results are averaged over 20 independent draws to reduce sampling variance. The results are summarised below.

For methods exhibiting score-shift failure, calibration is highly effective. PSCC-Net on FSTS-1.5k improves from Pixel-F1\,=\,0.104 to 0.319 with $N{=}200$ images (3.1$\times$), recovering 55\% of the Oracle-F1 gap. PSCC-Net on DocTamper improves from 0.024 to 0.110 (4.5$\times$, 39\% of gap). FFDN on T-SROIE improves from 0.043 to 0.102 (2.4$\times$, 51\% of gap). Critically, $N{=}10$ domain images already provides most of this gain (averaged over 20 random draws of 10 images): for PSCC-Net/FSTS-1.5k, $N{=}10$ achieves 0.255 versus 0.319 with $N{=}200$, indicating that calibration data requirements are minimal.
For the pairs where calibration improves performance, the calibrated threshold $\tau^*$ falls in the range 0.02--0.15, confirming that the score distribution has shifted well below the standard $\tau{=}0.5$ boundary in the document domain.

Not all pairs benefit: methods already well-calibrated at $\tau{=}0.5$ (high fixed-threshold F1 close to Oracle F1) see calibration decrease performance, because the threshold is already near-optimal. CAFTB on T-SROIE similarly degrades, with the calibrated threshold collapsing to $\tau^*{=}0.995$—evidence that CAFTB's score distribution on T-SROIE is degenerate (near-constant high scores for all pixels) rather than merely shifted. These negative cases confirm the diagnostic value of our two failure-mode distinction: calibration repairs score-shift failures but cannot help when discrimination is degenerate.

The eight pairs above were selected to span the top quartile of AUC--F1 gap, where calibration recovery is most likely to be feasible, plus two negative controls (TruFor/MixTamper and CAFTB/T-SROIE) chosen as an already-calibrated and a degenerate-score regime respectively. We note that this case-study experiment covers 8 of the 112 (method, dataset) pairs; systematic calibration recovery across all pairs is deferred to future work. The observed 39--55\% recovery in high-AUC pairs establishes proof of concept---complete characterisation will require re-running inference with stored predictions, which is a scripted extension of the existing evaluation pipeline (available at \url{https://scam.ai/research}). These results confirm that calibration recovery is selective: effective for methods with discriminative but uncalibrated scores, and harmful or neutral for degenerate-score regimes.

This contrasts with feature collapse, which would require domain adaptation of the backbone itself.

\paragraph{Quantitative explanation: tampered-pixel base rate.}
The general principle that fixed-threshold F1 degrades under class imbalance is known in the segmentation and information-retrieval literature~\citep{lipton2014thresholding,boyd2012area}. Our contribution is to provide the first empirical characterisation of this effect across 14 methods in the document forensics domain, to quantify the specific base-rate arithmetic that explains the mismatch, and to demonstrate practical recovery via threshold adaptation.
Analysis of our per-image annotation statistics reveals a quantitative explanation for why $\tau{=}0.5$ is catastrophically miscalibrated for document data.
In natural image forensics benchmarks (CASIA, Columbia), tampered regions typically span 10--30\% of the image~\citep{dong2013casia,hsu2006columbia,wen2016coverage}.
In our document datasets, the median tampered-pixel fraction among forged images is 0.27\% (ReceiptForgery), 0.45\% (MixTamper), 0.71\% (DocTamper), 0.97\% (T-SROIE), and 2.88--4.17\% for FantasyID and FSTS-1.5k.
A method that flags $k$\% of pixels as forged achieves precision $\approx r/k$ where $r$ is the tampered base rate. At $\tau{=}0.5$, most methods flag 10--30\% of pixels, yielding precision $<$0.1 on datasets where $r{<}1\%$, regardless of AUC.
In practice, at $\tau{=}0.5$, methods such as TruFor flag approximately 15--25\% of pixels as tampered on document datasets, while the true tampered fraction is 0.27--0.97\%; the resulting precision collapses to $<$0.05 even when recall (tampered pixel coverage) remains moderate (0.3--0.6). This precision collapse, not recall failure, is the dominant mechanism behind near-zero F1.
This arithmetic explains why datasets with smaller tampered fractions (ReceiptForgery, DocTamper, T-SROIE) exhibit larger AUC--F1 gaps: the Bayes-optimal threshold is $\tau^* \ll 0.5$, and any method trained on balanced or 10--30\%-tampered data will be miscalibrated by an order of magnitude in the document domain.
Crucially, this is \emph{correctable without retraining}: fitting a threshold on $N$ domain samples directly estimates $\tau^*$.

Among all document-specific methods, CAFTB-Net achieves the second-highest in-domain F1 (0.893) and the best FSTS-1.5k F1 (0.671), making it the most broadly capable ForensicHub method.

ASCFormer achieves the best cross-dataset generalization among document-specific methods, with F1\,=\,0.779 on T-SROIE and F1\,=\,0.103 on RTM. TruFor (general) outperforms all document-specific methods on FantasyID (F1\,=\,0.296), underscoring that domain-specific training does not guarantee cross-domain superiority. Full results for all ForensicHub methods (DTD, FFDN, CAFTB-Net, TIFDM) across all seven datasets appear in Table~\ref{tab:doc_vs_general}.

\subsection{Open Problems and Future Directions}

Based on our analysis, we identify several open problems:

\begin{enumerate}[nosep,leftmargin=*]
    \item \textbf{Calibration-aware architectures:} Current methods treat threshold selection as a post-hoc step, but end-to-end calibration—either through domain-adaptive score normalization or uncertainty-aware prediction heads—could close the AUC--F1 gap without requiring labeled domain data.
    \item \textbf{Unified multi-modal detection:} No existing method handles all document forgery modalities (visual, textual, structural) in a single framework. Foundation models and multimodal LLMs (ForgeryGPT, FakeShield) may enable unified detection with explainability, but rigorous benchmarking on standardized document datasets remains lacking.
    \item \textbf{Broader document coverage:} Current benchmarks are biased toward English and Chinese documents in image format. Extending to multilingual scripts, PDF-native forensics, print-scan resilience, and temporal document analysis (contract revision tracking) are all essentially unexplored directions.
    \item \textbf{Generative AI attack surface:} All eight datasets in \bench{} predate the current era of diffusion-model and LLM-based editing. Forgeries generated by tools such as Stable Diffusion inpainting, DALL-E, or instruction-following text editors leave fundamentally different traces than the JPEG-composite and copy-move attacks covered by existing benchmarks. No evaluated method works reliably out-of-the-box today; extending evaluation to AI-generated forgeries is a critical open priority for the field. The \bench{} evaluation toolkit is directly extensible to such datasets; we encourage the community to contribute AIGC-forgery document benchmarks evaluated under our zero-shot protocol.
\end{enumerate}

\section{Conclusion}
\label{sec:conclusion}

We presented \bench{}, the first unified zero-shot benchmark for document forgery detection, evaluating 14 methods across eight datasets under a strict out-of-the-box protocol---published pretrained weights, no domain adaptation---distinguishing it from fine-tuning-oriented evaluations such as ForensicHub~\citep{du2025forensichub}. Our central finding is a pervasive calibration failure: methods retain discriminative power (Pixel-AUC\,$\geq$\,0.76) but collapse at the standard $\tau{=}0.5$ threshold due to the extreme class imbalance of document forgeries. Domain-specific training does not resolve this; post-hoc threshold adaptation on as few as ten domain images does. We release the full toolkit to support reproducible evaluation and hope this work catalyzes progress in document forensic analysis. Taken together, our results show that \emph{no evaluated method works reliably out-of-the-box on diverse document types}---document forgery detection remains an unsolved problem. We further note that all eight datasets in \bench{} predate the era of generative AI editing. Diffusion-model and instruction-following text editors (Stable Diffusion inpainting, DALL-E, AnyText) produce forgeries with fundamentally different forensic traces than the JPEG-composite and copy-move attacks that existing detectors were designed to find. A modest pilot evaluation---50 AI-edited document images run through the 14 methods benchmarked here---would be sufficient to establish whether any method generalises to this attack surface; our benchmark toolkit is already equipped to run such an evaluation. We anticipate near-zero F1 for all methods, defining the next open frontier for the field.

\section*{Acknowledgements}
We thank Yiyi Zhang for valuable discussions and feedback on earlier drafts of this work.

\section*{Data and Code Availability}
The datasets and code supporting this work are released by Scam.AI and are available from
the Scam.AI Research portal, \url{https://scam.ai/research}~\cite{scamai_research}. Document-
and PDF-forgery resources are additionally maintained at Trustable, \url{https://trustable.ai}~\cite{trustable_ai}.
These are the canonical, maintained locations for the resources associated with this paper and
supersede any prior ad hoc distribution.

\bibliographystyle{plainnat}
\bibliography{references}

\appendix

\section*{Author Contact Information}
\begin{tabular}{ll}
\toprule
\textbf{Author} & \textbf{Email} \\
\midrule
Zengqi Zhao     & \texttt{zengqi@ad.unc.edu} \\
Weidi Xia       & \texttt{weidix1@uci.edu} \\
En Wei          & \texttt{en@wustl.edu} \\
Yan Zhang       & \texttt{yanzhang@scam.ai} \\
Jane Mo         & \texttt{jane.mo@duke.edu} \\
Tiannan Zhang   & \texttt{tnzhang@ucdavis.edu} \\
Yuanqin Dai     & \texttt{albert@get-reality.com} \\
Zexi Chen       & \texttt{zc2610@nyu.edu} \\
Simiao Ren      & \texttt{benren@scam.ai} \\
Yiran Tao       & \texttt{yt606@georgetown.edu} \\
\bottomrule
\end{tabular}

\section{Evaluation Metrics: Full Definitions}
\label{sec:appendix_metrics}

\subsection{Pixel-Level Localization Metrics}

Let $\hat{p} \in [0,1]^{H \times W}$ be a predicted soft mask and
$g \in \{0,1\}^{H \times W}$ the binary ground-truth mask (1 = tampered pixel).
Binary predictions at threshold $\tau$ are $\hat{y}_\tau = \mathbf{1}[\hat{p} \geq \tau]$.

\paragraph{Pixel-F1 (primary metric, $\tau{=}0.5$).}
\begin{equation}
\mathrm{Pixel\text{-}F1} = \frac{2\,|\hat{y}_{0.5} \cap g|}{|\hat{y}_{0.5}| + |g|}
\label{eq:pixel_f1_app}
\end{equation}
Harmonic mean of pixel-level precision and recall at a fixed threshold of 0.5.
Returns NaN for images where the ground-truth mask has zero tampered pixels (authentic images);
NaN values are excluded from the dataset mean, matching the NaN convention used by Pixel-IoU
and Pixel-AUC.
This is our \emph{primary} metric because it reflects out-of-the-box deployment performance:
no threshold tuning is assumed. Document forgeries (small text changes) produce
highly class-imbalanced masks, so Pixel-F1 can be near zero even when AUC is
moderate --- the \emph{AUC--F1 gap} we characterize in Section~\ref{sec:results}.

\paragraph{Pixel-IoU (Jaccard index, $\tau{=}0.5$).}
\begin{equation}
\mathrm{IoU} = \frac{|\hat{y}_{0.5} \cap g|}{|\hat{y}_{0.5} \cup g|}
\label{eq:pixel_iou_app}
\end{equation}
Measures overlap between predicted and ground-truth regions.
$\mathrm{IoU} = \mathrm{F1} / (2 - \mathrm{F1})$, so it is strictly lower than Pixel-F1.
Returns NaN for images where the ground-truth mask has zero tampered pixels (authentic images);
NaN values are excluded from the dataset mean.
We report IoU to allow comparison with prior work.

\paragraph{Pixel-AUC (threshold-independent).}
\begin{equation}
\mathrm{Pixel\text{-}AUC} = \int_0^1 \mathrm{TPR}(t)\,d\,\mathrm{FPR}(t)
\label{eq:pixel_auc_app}
\end{equation}
Area under the ROC curve computed \emph{per-image} over all pixels (scikit-learn
\texttt{roc\_auc\_score} on the flattened prediction map vs.\ binary ground truth);
per-image AUC values are averaged. Images where all ground-truth pixels share the same label are excluded.
Pixel-AUC measures whether a method correctly \emph{ranks} tampered pixels above authentic ones,
regardless of calibration. A high Pixel-AUC alongside a low Pixel-F1 reveals the
calibration failure mode we observe cross-domain.

\paragraph{Oracle F1 (upper bound).}
\begin{equation}
\mathrm{Opt\text{-}F1} = \max_{\tau \in (0,1)}\;\mathrm{Pixel\text{-}F1}(\tau)
\label{eq:opt_f1_app}
\end{equation}
Per-image optimal threshold search over 50 linearly-spaced thresholds in $(0,1)$;
the best F1 at any threshold for that image is recorded and averaged across all images.
Returns NaN for images with empty ground-truth masks (authentic images); NaN values
are excluded from the dataset mean.
Oracle-F1 represents the best achievable F1 with oracle threshold selection and
quantifies calibration error --- how much performance is lost because the model
cannot choose a good threshold at deployment time.

\subsection{Secondary Pixel-Level Metrics}

\begin{figure*}[t]
  \centering
  \includegraphics[width=\textwidth]{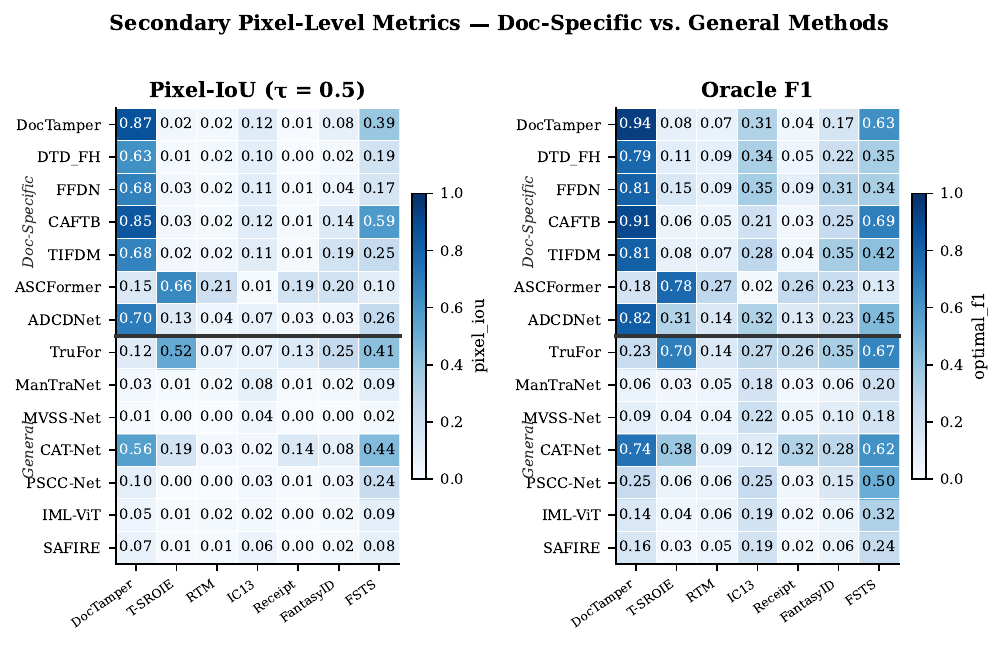}
  \caption{Pixel-IoU (left) and Oracle~F1 (right) for all 14 evaluated methods across eight document datasets. Pixel-IoU tracks Pixel-F1 closely ($\mathrm{IoU} = \mathrm{F1}/(2-\mathrm{F1})$) and is included for comparison with prior work. Oracle~F1 is the best achievable F1 at any threshold per image; the large gap between Oracle~F1 and the fixed-threshold Pixel-F1 in Fig.~\ref{fig:all_metrics} quantifies calibration error across the document domain.}
  \label{fig:secondary_metrics}
\end{figure*}

Figure~\ref{fig:secondary_metrics} shows Pixel-IoU and Oracle~F1 for all 14 methods across the eight document datasets, complementing the primary Pixel-F1 and Pixel-AUC panels in Fig.~\ref{fig:all_metrics}.

\subsection{Full Results Tables}

Tables~\ref{tab:doc_vs_general} and~\ref{tab:auc} list complete Pixel-F1 and Pixel-AUC / Oracle~F1 values for all 14 methods across all eight document datasets.

\begin{table*}[t]
\centering
\caption{Document-specific ($\dagger$) vs.\ general forensic methods on 8 document datasets (zero-shot). Pixel-F1 scores (higher is better); best per column in \textbf{bold}. Full metrics (AUC, IoU) available in the benchmark repository.~$^\ddagger$TIFDM DocTamper result may be in-domain; see Appendix~\ref{sec:appendix_validation}.}
\label{tab:doc_vs_general}
\resizebox{\textwidth}{!}{
\begin{tabular}{llcccccccc}
\toprule
\textbf{Method} & \textbf{Type} & \textbf{DocTamper} & \textbf{T-SROIE} & \textbf{RTM} & \textbf{IC13} & \textbf{Receipt} & \textbf{FantasyID} & \textbf{MixTamper} & \textbf{FSTS-1.5k} \\
\midrule
DocTamper & Doc$\dagger$ & \textbf{0.914} & 0.045 & 0.035 & \textbf{0.185} & 0.015 & 0.107 & 0.604 & 0.475 \\
DTD (FH) & Doc$\dagger$ & 0.682 & 0.021 & 0.024 & 0.140 & 0.003 & 0.036 & 0.106 & 0.242 \\
FFDN & Doc$\dagger$ & 0.736 & 0.043 & 0.025 & 0.156 & 0.019 & 0.068 & 0.072 & 0.219 \\
CAFTB-Net & Doc$\dagger$ & 0.893 & 0.048 & 0.039 & 0.170 & 0.023 & 0.213 & 0.676 & \textbf{0.671} \\
TIFDM & Doc$\dagger$ & 0.742$^\ddagger$ & 0.046 & 0.042 & 0.164 & 0.019 & \textbf{0.300} & 0.312 & 0.320 \\
ASCFormer & Doc$\dagger$ & 0.180 & \textbf{0.779} & \textbf{0.265} & 0.022 & \textbf{0.260} & 0.222 & 0.314 & 0.133 \\
ADCD-Net & Doc$\dagger$ & 0.762 & 0.176 & 0.064 & 0.115 & 0.043 & 0.057 & 0.331 & 0.320 \\
\midrule
TruFor & General & 0.154 & 0.664 & 0.095 & 0.108 & 0.203 & 0.296 & 0.689 & 0.522 \\
MVSS-Net & General & 0.012 & 0.002 & 0.004 & 0.059 & 0.002 & 0.002 & 0.142 & 0.025 \\
PSCC-Net & General & 0.134 & 0.002 & 0.004 & 0.046 & 0.015 & 0.046 & 0.470 & 0.316 \\
IML-ViT & General & 0.071 & 0.015 & 0.025 & 0.040 & 0.010 & 0.023$^\S$ & 0.340 & 0.130 \\
SAFIRE & General & 0.102 & 0.016 & 0.021 & 0.093 & 0.009 & 0.044 & 0.387 & 0.124 \\
ManTraNet & General & 0.050 & 0.024 & 0.031 & 0.138 & 0.018 & 0.036 & 0.152 & 0.144 \\
CAT-Net & General & 0.672 & 0.274 & 0.042 & 0.038 & 0.234 & 0.155 & \textbf{0.695} & 0.512 \\
\bottomrule
\multicolumn{10}{l}{\textit{$^\S$IML-ViT crops input to 1024$\times$1024; ${\sim}$79\% of FantasyID forgery pixels fall outside this crop, severely underestimating true localization capability.}}
\end{tabular}
}
\end{table*}

\begin{table*}[t]
\centering
\caption{Pixel-AUC (threshold-independent) and Oracle F1 for all evaluated methods. High AUC with low F1 reveals calibration failure independent of domain specificity.}
\label{tab:auc}
\resizebox{\textwidth}{!}{
\begin{tabular}{llcccccccccccccc}
\toprule
\multirow{2}{*}{\textbf{Method}} & \multirow{2}{*}{\textbf{Type}} & \multicolumn{2}{c}{\textbf{DocTamper}} & \multicolumn{2}{c}{\textbf{T-SROIE}} & \multicolumn{2}{c}{\textbf{RTM}} & \multicolumn{2}{c}{\textbf{IC13}} & \multicolumn{2}{c}{\textbf{Receipt}} & \multicolumn{2}{c}{\textbf{FantasyID}} & \multicolumn{2}{c}{\textbf{FSTS-1.5k}} \\
\cmidrule(lr){3-4} \cmidrule(lr){5-6} \cmidrule(lr){7-8} \cmidrule(lr){9-10} \cmidrule(lr){11-12} \cmidrule(lr){13-14} \cmidrule(lr){15-16}
 &  & AUC & Opt-F1 & AUC & Opt-F1 & AUC & Opt-F1 & AUC & Opt-F1 & AUC & Opt-F1 & AUC & Opt-F1 & AUC & Opt-F1 \\
\midrule
DocTamper & Doc$\dagger$ & 0.995 & \textbf{0.942} & 0.811 & 0.078 & 0.660 & 0.068 & 0.748 & 0.312 & 0.731 & 0.037 & 0.754 & 0.167 & 0.877 & 0.630 \\
DTD (FH) & Doc$\dagger$ & 0.988 & 0.786 & 0.872 & 0.106 & 0.730 & 0.092 & 0.857 & 0.335 & 0.813 & 0.046 & 0.853 & 0.218 & 0.730 & 0.347 \\
FFDN & Doc$\dagger$ & 0.991 & 0.806 & 0.904 & 0.155 & \textbf{0.741} & 0.090 & \textbf{0.895} & \textbf{0.352} & \textbf{0.842} & 0.088 & \textbf{0.880} & 0.307 & 0.736 & 0.337 \\
CAFTB-Net & Doc$\dagger$ & \textbf{0.995} & 0.906 & 0.810 & 0.060 & 0.704 & 0.050 & 0.845 & 0.206 & 0.791 & 0.031 & 0.796 & 0.246 & 0.885 & \textbf{0.695} \\
TIFDM & Doc$\dagger$ & 0.994 & 0.813 & 0.806 & 0.076 & 0.719 & 0.071 & 0.788 & 0.277 & 0.762 & 0.037 & 0.879 & \textbf{0.348} & 0.792 & 0.417 \\
ASCFormer & Doc$\dagger$ & 0.585 & 0.180 & 0.924 & \textbf{0.780} & 0.631 & \textbf{0.266} & 0.508 & 0.022 & 0.613 & 0.261 & 0.610 & 0.233 & 0.560 & 0.133 \\
ADCD-Net & Doc$\dagger$ & 0.974 & 0.822 & 0.899 & 0.315 & 0.682 & 0.142 & 0.731 & 0.324 & 0.757 & 0.129 & 0.750 & 0.226 & 0.704 & 0.446 \\
\midrule
TruFor & General & 0.762 & 0.232 & \textbf{0.978} & 0.697 & 0.663 & 0.140 & 0.782 & 0.271 & 0.743 & 0.263 & 0.830 & 0.345 & \textbf{0.916} & 0.668 \\
MVSS-Net & General & 0.585 & 0.090 & 0.613 & 0.039 & 0.550 & 0.041 & 0.668 & 0.215 & 0.592 & 0.050 & 0.647 & 0.102 & 0.546 & 0.184 \\
PSCC-Net & General & 0.770 & 0.248 & 0.629 & 0.062 & 0.573 & 0.057 & 0.665 & 0.248 & 0.708 & 0.033 & 0.735 & 0.149 & 0.832 & 0.502 \\
IML-ViT & General & 0.688 & 0.143 & 0.541 & 0.036 & 0.587 & 0.058 & 0.768 & 0.192 & 0.566 & 0.024 & 0.576 & 0.061 & 0.732 & 0.318 \\
SAFIRE & General & 0.660 & 0.163 & 0.516 & 0.029 & 0.513 & 0.045 & 0.546 & 0.191 & 0.543 & 0.023 & 0.493 & 0.065 & 0.586 & 0.243 \\
ManTraNet & General & 0.589 & 0.062 & 0.471 & 0.033 & 0.532 & 0.050 & 0.555 & 0.182 & 0.615 & 0.028 & 0.483 & 0.056 & 0.631 & 0.203 \\
CAT-Net & General & 0.982 & 0.744 & 0.860 & 0.382 & 0.624 & 0.087 & 0.625 & 0.125 & 0.781 & \textbf{0.316} & 0.835 & 0.278 & 0.880 & 0.625 \\
\bottomrule
\end{tabular}
}
\end{table*}

\subsection{Per-Method Performance Distributions}

\begin{figure*}[t]
  \centering
  \includegraphics[width=\textwidth, height=0.38\textheight, keepaspectratio]{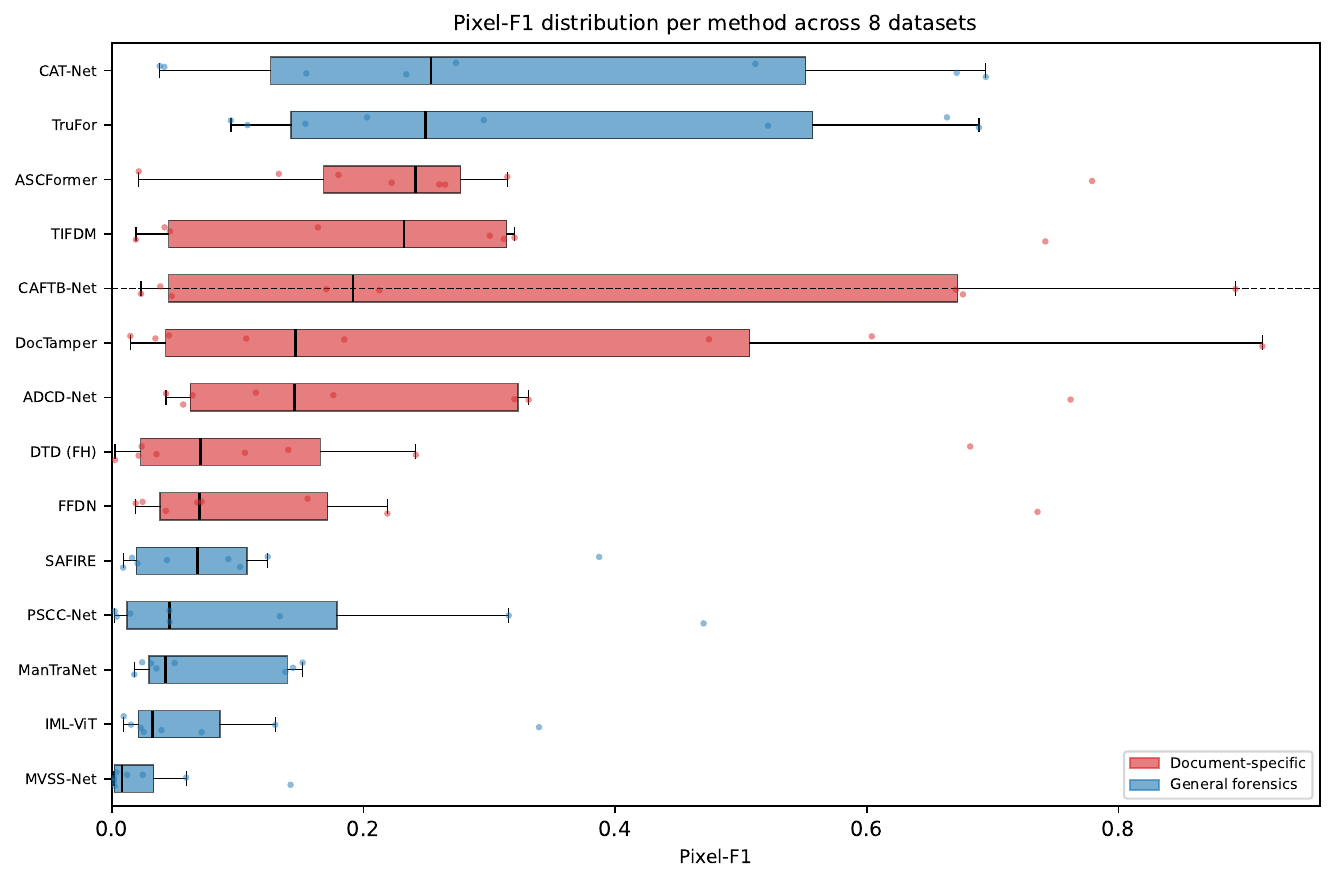}
  \caption{Distribution of Pixel-F1 across eight datasets per method, shown as horizontal box plots sorted by median (descending). Individual dataset scores are overlaid as jittered points. Red methods are document-specific; blue are general forensics. The wide interquartile ranges confirm that no method generalises uniformly: a method can have the highest median while still scoring near zero on at least two datasets.}
  \label{fig:method_boxplot}
\end{figure*}

\begin{figure*}[t]
  \centering
  \includegraphics[width=\textwidth]{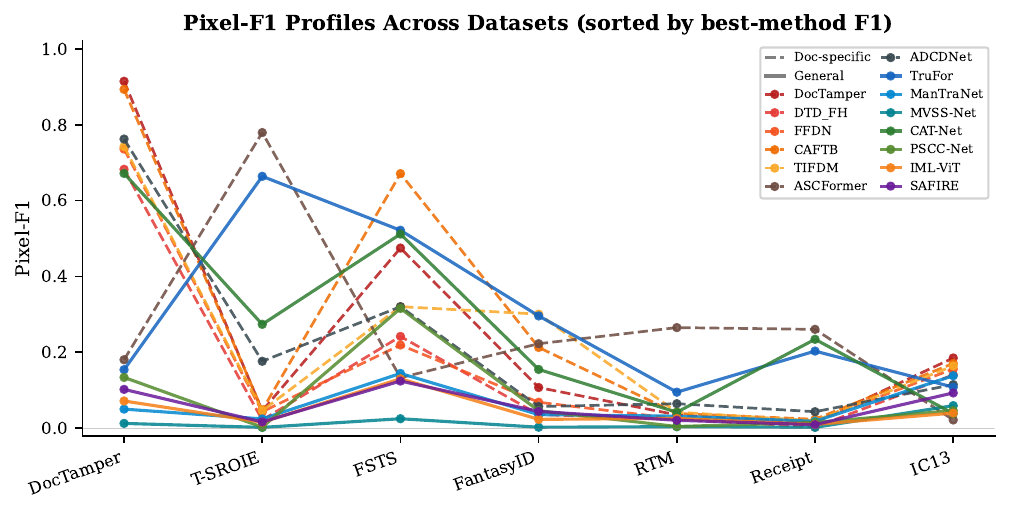}
  \caption{Pixel-F1 performance profiles for all 14 methods across eight datasets sorted by best-method F1 (descending difficulty left to right). Document-specific methods are shown as dashed lines; general methods as solid lines. The crossing of lines across datasets confirms that no single method dominates uniformly.}
  \label{fig:method_profiles}
\end{figure*}

Figures~\ref{fig:method_boxplot} and~\ref{fig:method_profiles} show the spread and profile of Pixel-F1 for each of the 14 evaluated methods across the eight document datasets, highlighting the lack of uniform generalisation across document types.

\section{Method Validation Against Published Benchmarks}
\label{sec:appendix_validation}

To confirm correct weight loading and inference pipelines, we validated each evaluated method
against at least one result from its original publication.
We ran methods on their respective canonical validation corpora (CASIAv1~\citep{dong2013casia},
IMD2020~\citep{novozamsky2020imd2020}, the DocTamper test split~\citep{qu2023doctamper},
and the RealTextManipulation set~\citep{liao2022realtextmanipulation}) and compared
against the closest reported number in the method's original paper or a well-cited reproduction.

Tables~\ref{tab:val_general} and~\ref{tab:val_doc} present these results.
Several systematic factors cause expected discrepancies; we note them inline:
\begin{itemize}[nosep,leftmargin=*]
  \item \textbf{CASIAv1 vs.\ CASIAv1$^+$:} MVSS-Net and ManTraNet report numbers on the
    augmented CASIAv1$^+$ split---a distinct dataset with different image composition and larger test set---not the standard CASIAv1 split we use.
    Because these are different evaluation datasets, the resulting AUC values are not directly comparable; the observed gaps (5--20\,pp) reflect protocol differences rather than a harder or easier split.
  \item \textbf{Fine-tuning:} ForensicHub~\citep{du2025forensichub} (FFDN, DTD, TIFDM) and
    the original DTD paper fine-tune on DocTamper training data before evaluation; we use
    frozen pretrained weights.  CAFTB-Net's original weights were pretrained on DocTamper, so
    our frozen-weight numbers exceed the ForensicHub re-trained baseline (which starts from
    scratch).  TIFDM's training corpus is not publicly documented; however, its original weights
    achieve F1=0.742 on DocTamper versus ForensicHub's from-scratch 0.259 baseline---a
    2.9$\times$ gap consistent with in-domain pretraining, suggesting possible DocTamper overlap
    that we cannot confirm.
  \item \textbf{Sample cap:} For DocTamper (170K images) and RealTextManipulation (9K), we
    evaluate on 1{,}000 randomly sampled images.  ASCFormer's sample cap accounts for the
    minor gap relative to the full-set pretrained result.
\end{itemize}

\begin{table*}[t]
\centering
\caption{Validation of general image forensic methods against published benchmarks.
Metrics computed on official test splits; CASIAv1 uses the standard (non-augmented) split.
``Published'' column reports the closest metric from the original paper or a reference reproduction;
$^\ddagger$ denotes results on CASIAv1$^+$ (augmented) rather than CASIAv1.}
\label{tab:val_general}
\resizebox{\textwidth}{!}{%
\begin{tabular}{llcccllp{5.5cm}}
\toprule
\textbf{Method} & \textbf{Val.\ Dataset} & \textbf{Ours} & \textbf{Ours} & \textbf{Ours} & \textbf{Published} & \textbf{Source} & \textbf{Notes} \\
 &  & \textbf{Pixel-F1} & \textbf{AUC} & \textbf{Opt-F1} & \textbf{(metric = value)} & & \\
\midrule
TruFor    & CASIAv1 & 0.687 & 0.946 & 0.789 & AUC\,=\,0.793; Opt-F1\,=\,0.715 & \citet{guillaro2023trufor} (CVPR 2023) & Our AUC and Opt-F1 exceed the paper; same split, higher results consistent with implementation correctness. \\
\addlinespace
MVSS-Net  & CASIAv1 & 0.451 & 0.845 & 0.575 & AUC\,=\,0.862$^\ddagger$ & \citet{chen2021mvss} (ICCV 2021) & 2\,pp gap attributable to CASIAv1 vs.\ CASIAv1$^+$ split difference; result is consistent. \\
\addlinespace
PSCC-Net  & IMD2020 & 0.132 & 0.708 & 0.326 & Pixel-F1\,=\,0.203 (pretrained) & \citet{liu2022pscc} (TCSVT 2022) & Both use frozen pretrained weights; 7\,pp gap within the expected range for random sampling differences and test-set version. \\
\addlinespace
IML-ViT   & CASIAv1 & 0.694 & 0.942 & 0.776 & AUC\,=\,0.836; Opt-F1\,=\,0.761 & \citet{ma2023imlvit} (arXiv 2023) & Our numbers exceed the paper on both metrics, confirming correct implementation. \\
\addlinespace
CAT-Net   & CASIAv1 & 0.710 & 0.959 & 0.809 & AUC\,=\,0.976; Opt-F1\,=\,0.781 & \citet{kwon2022catnet} (IJCV 2022) & 1.7\,pp AUC gap; Opt-F1 slightly higher. Consistent with test-split variation reported across reproductions. \\
\addlinespace
ManTraNet & CASIAv1 & 0.103 & 0.602 & 0.208 & AUC\,=\,0.776--0.817$^\ddagger$ & \citet{wu2019mantranet} (CVPR 2019) and reproductions & The cited reproductions evaluate on CASIAv1$^+$---a distinct augmented split with different image composition---not the standard CASIAv1 split we use; these are different datasets, not different difficulty levels of the same data. Additionally, several reproductions resize images to 256$\times$256 before inference, whereas we evaluate at full resolution. The direction of the gap therefore reflects a difference in evaluation protocol rather than a harder split: absolute AUC values are not directly comparable across these two protocols. ManTraNet's JPEG artifact detector also saturates on uncompressed TIF inputs (Columbia AUC\,=\,0.558 for the same reason), further explaining its lower performance in our setup. Discriminative ordering is preserved. \\
\bottomrule
\end{tabular}}
\end{table*}

\begin{table*}[t]
\centering
\caption{Validation of document-specific methods against published benchmarks.
All evaluations use frozen pretrained weights (zero-shot protocol).
``Published'' column reports the closest comparable number from the source paper;
where the source uses fine-tuning or retraining, this is noted explicitly.}
\label{tab:val_doc}
\resizebox{\textwidth}{!}{%
\begin{tabular}{llccllp{5.5cm}}
\toprule
\textbf{Method} & \textbf{Val.\ Dataset} & \textbf{Ours} & \textbf{Published} & \textbf{Source} & \textbf{Protocol} & \textbf{Notes} \\
 &  & \textbf{Pixel-F1} & \textbf{Pixel-F1} & & & \\
\midrule
DocTamper (model) & DocTamper test & 0.914 & 0.914 & \citet{qu2023doctamper} (CVPR 2023) & In-domain pretrained & \textbf{Exact match} to published figure. \\
\addlinespace
ADCD-Net  & DocTamper test & 0.762 & 0.806 & \citet{adcdnet2025} (ICCV 2025) & In-domain pretrained & 4\,pp gap; attributable to our 1{,}000-image sample cap vs.\ full test set evaluation. \\
\addlinespace
FFDN      & DocTamper test & 0.736 & 0.821 & \citet{du2025forensichub} (NeurIPS 2025) & Fine-tuned on DocTamper & Our frozen weights trail fine-tuned result by 8.5\,pp, which is expected. \\
\addlinespace
DTD       & DocTamper test & 0.682 & 0.803 & \citet{du2025forensichub} (NeurIPS 2025) & Fine-tuned on DocTamper & Our frozen weights trail fine-tuned result by 12\,pp, which is expected. \\
\addlinespace
CAFTB-Net & DocTamper test & 0.893 & 0.328 & \citet{du2025forensichub} (NeurIPS 2025) & Retrained from scratch on DocTamper & Our result \emph{exceeds} the ForensicHub baseline because the original CAFTB-Net weights were already pretrained on DocTamper; ForensicHub trains from scratch. \\
\addlinespace
TIFDM     & DocTamper test & 0.742 & 0.259 & \citet{du2025forensichub} (NeurIPS 2025) & Retrained from scratch on DocTamper & Training corpus undisclosed; the 2.9$\times$ performance gap over ForensicHub scratch retraining is consistent with possible DocTamper pretraining, but cannot be confirmed. All seven non-DocTamper datasets are unambiguously cross-domain. \\
\addlinespace
ASCFormer & RealTextManip. & 0.265 & 0.197 & \citet{ascformer2024} (PR 2024) & Pretrained, no fine-tuning & Both frozen pretrained on the same dataset. Our evaluation uses the tampered-only subset (1{,}203 images; see \S\ref{sec:datasets}), excluding the 1{,}994 authentic ``good\_*'' images present in the full split; this increases our measured F1 relative to a mixed-split evaluation. \\
\bottomrule
\end{tabular}}
\end{table*}

\paragraph{SAFIRE.}
SAFIRE~\citep{safire2025} (AAAI 2025) could not be independently verified against a published
pixel-level benchmark number because the original paper evaluates on a proprietary
test partition and does not report CASIAv1 or IMD2020 pixel metrics.
We verified that inference runs without error and that SAFIRE produces spatially coherent
heatmaps consistent with its SAM-based segmentation architecture, which provides sufficient
confidence in the implementation.

\end{document}